%% file: main.tex
\pdfoutput=1
\PassOptionsToPackage{table}{xcolor}
\documentclass{article}

\usepackage{iclr2027_conference,times}
\iclrfinalcopy

\usepackage[utf8]{inputenc}
\usepackage[T1]{fontenc}
\usepackage{amsmath,amssymb}
\usepackage{graphicx}
\usepackage{booktabs}
\usepackage{multirow}
\usepackage{float}
\usepackage{placeins}
\floatstyle{ruled}
\newfloat{algorithm}{tbp}{loa}
\floatname{algorithm}{Algorithm}
\usepackage{xcolor}
\usepackage{hyperref}
\usepackage{url}

\newif\ifrebuttal \rebuttalfalse
\newcommand{\rev}[1]{\ifrebuttal{\color{red}#1}\else#1\fi}
\newcommand{\revon}{\ifrebuttal\color{red}\fi}  


\title{GuidanceNFT: NFT-Based Reward Fine-Tuning of Guided Diffusion Models}
\author{Jiayin Liu, Zeyu Michael Li, Bingshuo Qian, Xiang Cheng\\
Duke University\\
\texttt{\{jiayin.liu2, zeyu.li030, bingshuo.qian, xiang.cheng\}@duke.edu}}

\begin{document}

\maketitle
\lhead{Preprint.}

\input{abstract}

\input{introduction}
\input{background}
\input{method}
\input{experiments}

\input{related_work}
\input{conclusion}
\input{statements}

\bibliographystyle{iclr2027_conference}
\bibliography{references}

\clearpage
\appendix
\input{appendix}

\end{document}

%% file: abstract.tex
\begin{abstract}
Guidance and reward fine-tuning improve diffusion generation in
complementary ways. Combining the two, however, can undermine
both benefits: applying classifier-free guidance after DiffusionNFT
(NFT) can reduce the optimized reward, while incorporating guidance
during NFT fine-tuning may cause collapse. We identify a mismatch in NFT's regression center as a source
of this instability. The center must
match the conditional mean velocity obtained by re-noising generated samples;
using a mismatched center introduces a bias that can rotate
or reverse the reward-driven update. We propose GuidanceNFT,
which learns this center online by flow matching on guided samples and
corrects both NFT regression terms while preserving the policy's
reward-driven displacement. The method supports explicit
and intrinsic guidance without additional inference passes. On SD3.5-M,
GuidanceNFT improves the optimized reward over standard NFT across three
training objectives and outperforms NFT on all seven evaluation metrics
for GenEval and OCR specialization. In particular, OCR reward increases from 0.971 to 0.975 while held-out
GenEval success rises from 0.035 to 0.488, indicating improved prompt
fidelity beyond the training objective.
Across ImageNet generation, OpenWebText, and GSM8K reasoning,
GuidanceNFT improves generation quality and task performance under
intrinsic guidance. These results show that online center correction enables effective
reward fine-tuning of guided diffusion models.
\end{abstract}

%% file: introduction.tex
\section{Introduction}

\begin{figure}[b]
\centering
\includegraphics[width=\textwidth]{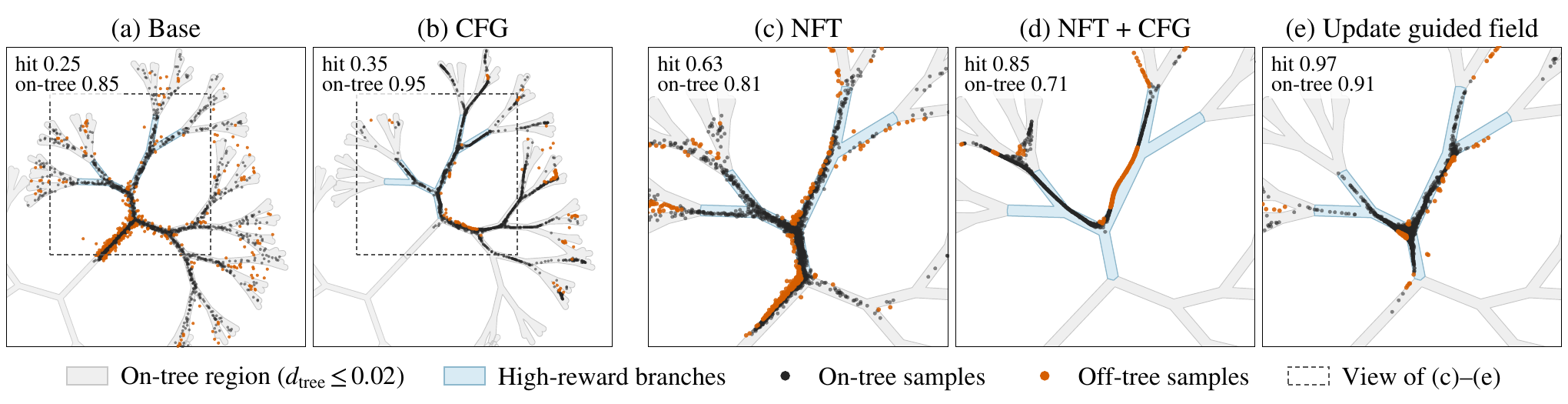}
\caption{{\revon\textbf{Intuition for updating a guided field.}
(a--b) Pretrained samples without and with CFG.
(c) NFT moves samples toward the rewarded branches, but (d) adding CFG
afterward pushes many samples off the tree.
(e) Updating the guided field better combines reward targeting with tree
fidelity. Panels (c--e) zoom into the dashed region; blue branches mark
the reward target and orange points mark off-tree samples.
Implementation details are in Appendix~\ref{app:toy-guidance-nft}.}}
\label{fig:toy-guidance-nft}
\end{figure}

Guidance and reward fine-tuning improve diffusion generation in
complementary ways. Classifier-free guidance (CFG) strengthens prompt
alignment by combining conditional and empty-condition predictions
\citep{ho2022cfg}, while reward fine-tuning adapts generation to a target
objective. DiffusionNFT (NFT)~\citep{zheng2026diffusionnft} performs this
adaptation through regression on positive and negative feedback,
without reward gradients or trajectory likelihoods. Combining guidance
with NFT could therefore improve task-specific performance while
retaining broader generation quality (Figure~\ref{fig:toy-guidance-nft}).

This broader quality is difficult to capture with a task reward alone.
For example, an OCR-specialized NFT model can render the requested
words while omitting the surrounding scene
(Figure~\ref{fig:sd35-learned3-ocr-qualitative}). Simply adding CFG after
fine-tuning does not resolve this limitation: in our SD3.5-M experiments,
it reduces the optimized reward for all three training objectives.
A natural alternative is to incorporate guidance during fine-tuning,
so that reward feedback reflects the samples used at deployment.
Yet standard NFT can then degrade or collapse, even when sampling and
training use the same guidance setting. This failure occurs with both
explicit CFG and \emph{intrinsic guidance}, where guidance is built into
the model's prediction (Figures~\ref{fig:explicit-guidance-diagnosis}
and~\ref{fig:intrinsic-guidance-dynamics}).

The purpose of this paper is to enable effective NFT fine-tuning of
guided diffusion models. We identify a mismatch in NFT's regression
objective that can undermine reward optimization under guidance.
NFT uses a shared reference prediction, or \emph{center}, to learn from
higher- and lower-reward samples. For these two regression terms to
recover the intended update, the center must match the conditional mean
velocity obtained by re-noising the generated samples. The guided
prediction used to generate those samples need not satisfy this
requirement, even when it combines exact component predictors
\citep{bradley2025predictor,choi2026rvm}. We derive how this mismatch
introduces a bias that can rotate or reverse the reward-driven update.
Our experiments link larger measured center gaps to faster collapse
and show that correcting the center sustains reward improvement.

Building on this analysis, we propose \emph{GuidanceNFT}, which estimates
the regression center separately from the policy being optimized.
An auxiliary predictor learns the center by flow matching on guided
samples and is refreshed as their distribution changes. We use its
difference from NFT's existing center to correct both regression terms,
holding the correction fixed during each policy update. NFT can then
optimize the guided model using a center fitted to its own samples.
The same construction supports explicit and intrinsic guidance, and
the auxiliary predictor is discarded at inference.

GuidanceNFT improves both task performance and generation quality.
On SD3.5-M, it exceeds NFT with and without CFG in the optimized reward
for all three training objectives. For GenEval and OCR specialization,
the gains extend to all seven evaluation metrics. In particular,
OCR reward increases from 0.971 to 0.975 while held-out GenEval success
rises from 0.035 to 0.488, illustrating improved prompt fidelity beyond
the training objective. With intrinsic guidance, GuidanceNFT improves
ImageNet generation, achieves a better perplexity--entropy tradeoff on
OpenWebText, and outperforms NFT's best retained checkpoint at every
reported Pass@$k$ on GSM8K.

Our main contributions are:
\begingroup
\setlength{\topsep}{2pt}
\setlength{\partopsep}{0pt}
\setlength{\itemsep}{2pt}
\setlength{\parsep}{0pt}
\setlength{\parskip}{0pt}
\begin{enumerate}
\item \textbf{A diagnosis of NFT instability under guidance.}
We identify a mismatch between NFT's reference prediction and the
distribution of samples produced under guidance.
We derive how this mismatch biases the reward-driven update and
provide empirical evidence linking it to performance degradation and
training collapse.
\item \textbf{An online correction for guided reward fine-tuning.}
We introduce GuidanceNFT, which learns the required regression center
from guided samples and corrects the NFT objective as the policy evolves.
The method supports explicit and intrinsic guidance without additional
inference passes.
\item \textbf{Improved task performance and overall generation quality.}
GuidanceNFT improves task rewards and overall generation quality on
SD3.5-M. It also enables NFT to improve models with intrinsic guidance,
including Model Guidance and self-conditioning CFG (SCCFG).
Experiments on ImageNet, OpenWebText, and GSM8K further demonstrate
performance gains across models, tasks, and guidance mechanisms.
\end{enumerate}
\endgroup

%% file: background.tex
\section{Background}
\label{sec:background}

\subsection{DiffusionNFT}

DiffusionNFT~\citep{zheng2026diffusionnft} learns from the contrast
between higher- and lower-reward samples. Given a rollout distribution
$q$ and processed reward weights $r\in[0,1]$, it defines positive and
negative populations $q^+\propto rq$ and $q^-\propto(1-r)q$.
Their conditional mean velocities provide the optimal regression
targets for the two populations. Rather than fitting these targets
with separate models, NFT parameterizes both predictions using a
trainable policy $v_\theta$ and a frozen old policy $v^{\rm old}$:
\begin{equation}
v_\theta^+=(1-\beta)v^{\rm old}+\beta v_\theta,\qquad
v_\theta^-=(1+\beta)v^{\rm old}-\beta v_\theta.
\label{eq:nft-implicit-predictions}
\end{equation}
These predictions move in opposite directions around $v^{\rm old}$,
with $\beta>0$ controlling their separation. Both are evaluated at the
same noisy sample $(x_t,c,t)$. Writing $v$ for its forward-process
velocity target, NFT minimizes
\begin{equation}
\mathcal L_{\rm NFT}(\theta)=\mathbb E\!\left[
r\|v_\theta^+-v\|^2+(1-r)\|v_\theta^--v\|^2
\right].
\label{eq:nft-objective}
\end{equation}
The shared parameterization turns positive and negative feedback into
a single policy update, using only generated endpoints and their
rewards. Appendix~\ref{app:nft-theory} specifies the forward process
and normalized populations.

\subsection{Classifier-Free and Built-in Guidance}
\label{sec:guidance-background}

\paragraph{Classifier-free guidance.}
CFG strengthens conditioning by extrapolating from the empty-condition
prediction $v_\varnothing$ toward the conditional prediction $v_c$
\citep{ho2022cfg}:
\begin{equation}
v_w=v_\varnothing+w(v_c-v_\varnothing)
   =w v_c+(1-w)v_\varnothing.
\label{eq:guided-field}
\end{equation}
Here $w=1$ gives conditional-only sampling. The guided field determines
the generated endpoints, but its intermediate predictions need not
equal the mean velocities obtained by re-noising those endpoints.
This distinction holds even for exact component
predictors~\citep{bradley2025predictor,chidambaram2024guidance}
and becomes central when using guided samples in NFT.

\paragraph{Model Guidance.}
Model Guidance (MG)~\citep{tang2025modelguidance} incorporates guidance
into the model's output, requiring one prediction per inference step. In velocity
coordinates, it trains $M_c$ toward
$v+\lambda\operatorname{sg}(M_c-M_\varnothing)$. For a fixed
$0\leq\lambda<1$, its ideal fixed point has the CFG form
\begin{equation}
M_c=w_{\rm eff}C_{\rm data}+(1-w_{\rm eff})M_\varnothing,
\qquad w_{\rm eff}=\frac{1}{1-\lambda},
\label{eq:mg-native-guidance}
\end{equation}
where $C_{\rm data}$ is the unweighted conditional mean velocity of
the pretraining data. The native prediction therefore represents a
guided field even without an external CFG operation.

\paragraph{Self-conditioning CFG (SCCFG).}
ELF~\citep{hu2026elf} learns guidance from its own clean-embedding
prediction. The same network produces a self-conditioned field
$S_\theta^w$ and a field $S_{\theta,\varnothing}^w$ with zero
self-conditioning input; both retain any task prompt. The
self-conditioning input comes from a detached initial prediction
during training and from the preceding sampling step at inference.
The training target incorporates guidance through
\begin{equation}
v_{\rm target}=v+\left(1-\frac1w\right)
\operatorname{sg}(S_\theta^w-S_{\theta,\varnothing}^w).
\label{eq:elf-native-guidance}
\end{equation}
Its corresponding CFG form is
$S_\theta^w=wB_\theta^w+(1-w)S_{\theta,\varnothing}^w$, where
$B_\theta^w$ is the algebraically reconstructed base prediction
(Appendix~\ref{app:intrinsic-center}).

Both MG and SCCFG thus expose NFT to a guided native prediction.
Setting external CFG to 1 preserves this built-in guidance, while
changing the internal scale changes the rollout distribution itself.
To update the original guided distribution with NFT, we therefore
need the center induced by its own re-noised endpoints.

%% file: method.tex
\section{GuidanceNFT}
\label{sec:method}
\label{sec:matched-center}

Applying NFT to guided rollouts requires its regression objective to
recover the reward-driven change in velocity. We first derive this
change and show how an inaccurate center distorts it. This identifies
both the quantity to estimate and the correction needed for guided
fine-tuning.

\subsection{Center Exactness Determines the NFT Update}

Fix the guided rollout law $q_{\rm CFG}(\cdot\mid c)$ and processed
reward weights $r\in[0,1]$ for one policy update. Here $q_{\rm CFG}$
also covers built-in guidance. At a re-noised state $z=(x_t,c,t)$,
let $\alpha=\mathbb E_{q_{\rm CFG}}[r\mid z]\in(0,1)$. The optimal
regression fields for the positive and negative populations are
\begin{equation}
v^+=\frac{\mathbb E_{q_{\rm CFG}}[rv\mid z]}{\alpha},
\qquad
v^-=\frac{\mathbb E_{q_{\rm CFG}}[(1-r)v\mid z]}{1-\alpha},
\qquad \Delta=v^+-v^-.
\label{eq:guided-reward-direction}
\end{equation}
Thus $\Delta$ separates the velocity associated with higher-reward
samples from that associated with lower-reward samples. Under Gaussian
noising, $v^+$ and $v^-$ are affine transforms of the corresponding
noised-population scores (Appendix~\ref{app:nft-theory}). NFT aims to
learn a displacement along this reward-improving direction without
fitting the two fields separately.

To determine when NFT recovers this direction, write the policy
displacement as $d=v_\theta-v^{\rm old}$ and center its implicit
predictions at a frozen field $C$. Standard NFT uses $C=v^{\rm old}$.
Conditioning its squared loss on $z$ gives, up to a constant in $d$,
\begin{equation}
\ell_z(C;d)
=\alpha\|C+\beta d-v^+\|^2
 +(1-\alpha)\|C-\beta d-v^-\|^2.
\label{eq:main-population-loss}
\end{equation}
Differentiating with respect to $d$ yields
$\beta d^*=\alpha(v^+-C)-(1-\alpha)(v^--C)$.
The positive term pulls toward $v^+$; the negative term pushes away
from $v^-$. To recover NFT's prescribed improvement from both branches,
their weighted residuals must satisfy
$\alpha(v^+-C)=(1-\alpha)(C-v^-)=\alpha(1-\alpha)\Delta$.
Solving either equality gives the unique matched center:
\begin{equation}
\bar v_{q_{\rm CFG}}
=\mathbb E_{q_{\rm CFG}}[v\mid z]
=\alpha v^++(1-\alpha)v^-.
\label{eq:matched-center}
\end{equation}
At this center the two residuals are
$v^+-\bar v_{q_{\rm CFG}}=(1-\alpha)\Delta$ and
$v^--\bar v_{q_{\rm CFG}}=-\alpha\Delta$, so both terms contribute
in the direction of $\Delta$.

\begin{figure}[t]
\centering
\includegraphics[width=\textwidth]{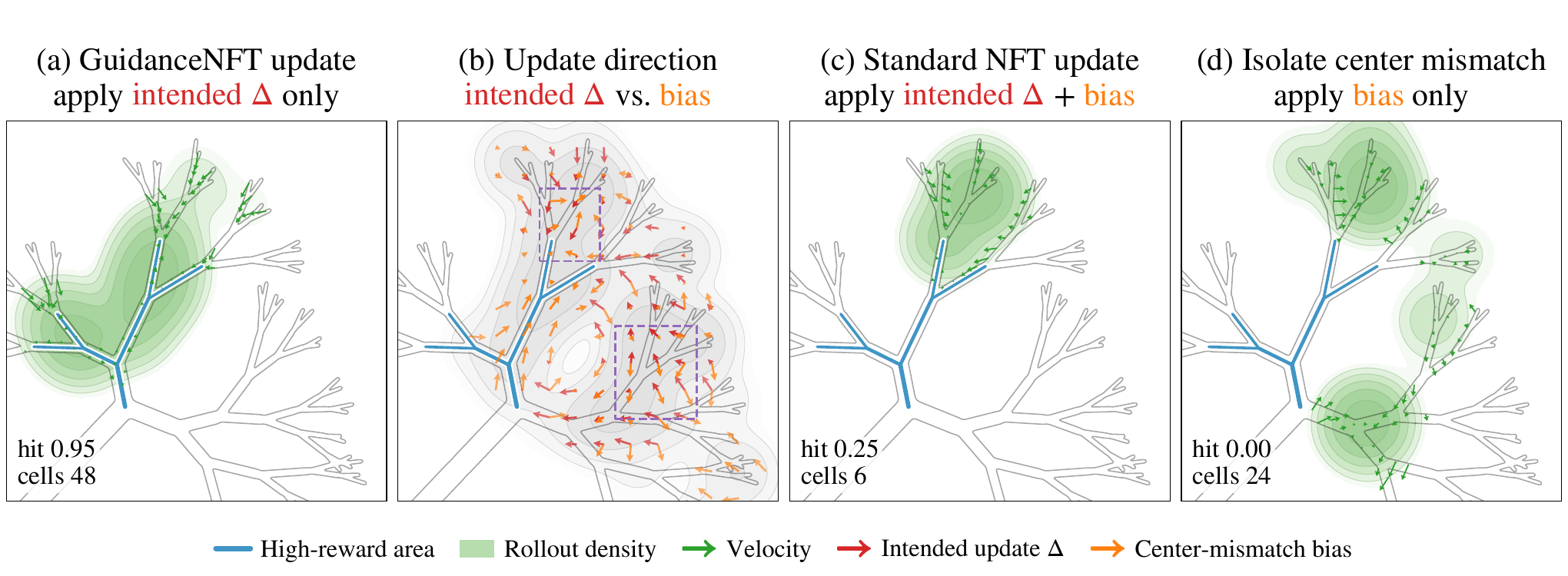}
\caption{{\revon\textbf{Center error distorts the reward-driven update.}
(b) On a fixed rollout, the center-induced bias has a magnitude comparable
to the intended displacement $\Delta$ and opposes it in large regions
(dashed boxes).
(a) Updating with $\Delta$ alone moves samples onto the rewarded branches,
whereas (c) adding the bias collapses them onto a small region.
(d) Applied alone for ten rounds, the bias moves samples away from the
reward target and distorts the velocity field, shown on the data
distribution. Details are in Appendix~\ref{app:toy-center-bias}.}}
\label{fig:toy-guidance-nft-center}
\end{figure}

For an arbitrary center, substituting
$C=\bar v_{q_{\rm CFG}}+e$ instead gives the exact decomposition
\begin{equation}
\boxed{\displaystyle
 d^*=\underbrace{[2\alpha(1-\alpha)/\beta]\Delta}_{\text{reward-driven displacement}}
     +\underbrace{[(1-2\alpha)/\beta]e}_{\text{center-induced bias}}.}
\label{eq:center-error}
\end{equation}
Thus the unique matched center recovers NFT's intended displacement
by making the two weighted branches contribute equally to it.
An error parallel to $\Delta$ can change its magnitude or reverse
its sign; an orthogonal component can rotate the update away from the
reward direction. Matching the center removes this bias across local
reward balances, preventing the objective from learning a spurious
correction alongside the reward signal.
\rev{Figure~\ref{fig:toy-guidance-nft-center} isolates this bias in a toy
model with a fixed rollout: applied alone, the center-induced term moves
samples away from the reward target, and added to the reward-driven
displacement it distorts the velocity field and lowers the reward
(details in Appendix~\ref{app:toy-center-bias}).} The identity holds
for the unrestricted optimum of the fixed-rollout surrogate;
Appendix~\ref{app:center-optimization} also derives the effect of center
error on the parameter gradient before convergence.

\subsection{Why Guidance Can Introduce a Center Gap}

A guided sampler generates endpoints by integrating a velocity field.
NFT then draws fresh noisy states from those endpoints. The sampler's
intermediate distributions need not coincide with the forward-noised
endpoint distributions, so its velocity need not supply NFT's center.
For the linear noising path $x_t=tx+(1-t)\epsilon$, this distinction
has the explicit form
\begin{equation}
\bar v_{q_{\rm CFG}}(x_t,c,t)
=\frac{x_t}{t}+\frac{1-t}{t}\nabla_{x_t}\log q_{{\rm CFG},t}(x_t\mid c),
\qquad 0<t<1,
\label{eq:main-center-score}
\end{equation}
where $q_{{\rm CFG},t}$ is obtained by re-noising the guided endpoints.
The relevant score is therefore determined by these endpoints and the
training noising process.

Even exact conditional and empty-condition predictors do not generally
make the analytic CFG field $v_w$ equal this center
\citep{bradley2025predictor,chidambaram2024guidance}.
Appendix~\ref{app:guided-score-mismatch} gives an exact Gaussian example
and shows why combining component scores need not preserve consistency
with forward noising. Built-in guidance has the same issue: a native
guided output need not equal the mean velocity of its re-noised
endpoints. Consequently, using the same guidance during sampling and
training leaves the bias in Equation~\ref{eq:center-error} unresolved.
Our experiments connect this gap to reward degradation and collapse
using separately fitted rollout references and paired held-out
flow-matching risks (Section~\ref{sec:explicit-experiments}).

\subsection{Estimating and Applying the Correction}

We correct the center online while retaining NFT's policy displacement.
Specifically, we add the same frozen correction $\mathcal C$ to both
implicit predictions:
\begin{equation}
\mathcal L_{\rm NFT}(\theta;\mathcal C)
=\mathbb E_{q_{\rm CFG}}\!\left[
r\|v_\theta^++\mathcal C-v\|^2
+(1-r)\|v_\theta^-+\mathcal C-v\|^2\right].
\label{eq:corrected-nft-objective}
\end{equation}
The ideal correction $\mathcal C^*=\bar v_{q_{\rm CFG}}-v^{\rm old}$
replaces the regression center with $\bar v_{q_{\rm CFG}}$ while leaving
$d=v_\theta-v^{\rm old}$ intact. This separation allows NFT to update a
guided prediction directly: the rollout distribution determines the
center, and reward feedback determines the displacement from the
current policy.

We estimate the center with an auxiliary predictor $v_\phi$ trained by
ordinary flow matching on guided endpoints:
\begin{equation}
\mathcal L_{\rm FM}(\phi)
=\mathbb E_{q_{\rm CFG}}\|v_\phi(x_t,c,t)-v\|^2,
\qquad
\mathcal C_\phi=\operatorname{sg}(v_\phi-v^{\rm old}).
\label{eq:center-fm}
\end{equation}
The population optimum is precisely $\bar v_{q_{\rm CFG}}$
\citep{lipman2023flowmatching}. We refresh $v_\phi$ as the rollout
population changes and freeze its snapshot within each NFT update.
The stop-gradient operator $\operatorname{sg}$ ensures that the NFT
loss updates only $v_\theta$. Its residual center bias is then governed
by $e=v_\phi-\bar v_{q_{\rm CFG}}$ in Equation~\ref{eq:center-error}.
Setting $\mathcal C=0$ recovers standard NFT; substituting the analytic
CFG field for $v_\phi$ gives the analytic-correction baseline.
The auxiliary predictor is discarded at inference, so correction adds
no sampler passes.

Algorithm~\ref{alg:guided-nft} summarizes the procedure. Exact center
matching requires the two objectives to share endpoint selection,
replay, masks, and conditioning
(Appendix~\ref{app:training-measure}). For MG and ELF, $v_\theta$ is the
native guided prediction. Our explicit-CFG experiments update the
conditional prediction within the guided sampler, as specified in
Appendix~\ref{app:experimental-protocol}.

%% file: experiments.tex
\section{Experiments}
\label{sec:experiments}

Our experiments show that center correction stabilizes guided NFT
and improves both the optimized reward and generation quality.
We first connect center mismatch to degradation under explicit CFG,
then test the correction with built-in guidance for image generation,
language modeling, and reasoning. Training protocols, uncertainty
estimates, and costs appear in Appendices~\ref{app:experimental-protocol}
and~\ref{app:center-diagnostics}.

\subsection{Explicit CFG: Diagnosing Degradation and Correcting the Center}
\label{sec:explicit-experiments}

\begin{figure}[!htbp]
\centering
\includegraphics[width=\linewidth]{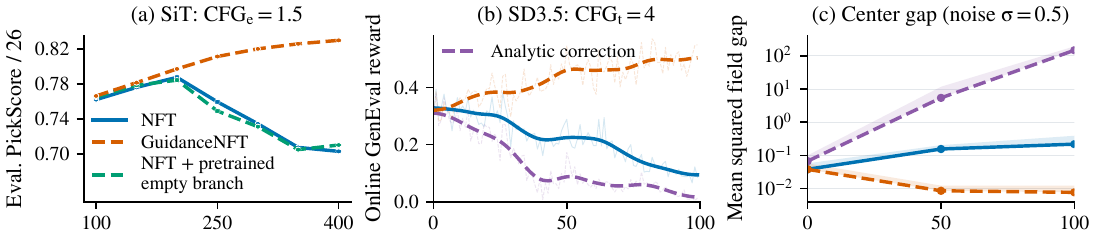}
\caption{\textbf{Correcting NFT's center prevents degradation under CFG.}
(a) NFT loses reward even after restoring the pretrained empty branch.
(b) Online center correction sustains reward improvement on SD3.5-M;
uncorrected and analytically corrected NFT collapse.
(c) Faster collapse accompanies larger gaps to separately fitted
rollout references. Gaps are measured at $\sigma=0.5$ using each run's
FM100 reference. \rev{Horizontal axes show training updates.}
Bars in (a) show evaluation-seed std; bands in (c)
span the mean to one prompt std above it.
Appendix~\ref{app:center-diagnostics} gives the diagnostic protocol.}
\label{fig:explicit-guidance-diagnosis}
\label{fig:sd35-guidance-failures}
\end{figure}

Standard NFT can lose its initial reward gains when deployed with
guidance. On ImageNet~\citep{russakovsky2015imagenet}, the reward of
SiT-XL/2~\citep{ma2024sit} under CFG peaks and then declines as NFT
training continues, whereas GuidanceNFT keeps improving
(Figure~\ref{fig:explicit-guidance-diagnosis}(a)). Restoring the
pretrained unconditional prediction, following
\citet{phunyaphibarn2026unconditional}, leaves this decline largely
intact. The failure therefore extends beyond damage to the
unconditional prior and motivates examining the center used by the
NFT objective.

Center correction resolves the more severe failure that arises when
NFT trains directly on guided samples. With CFG-4 rollouts from
SD3.5-M~\citep{stabilityai2024sd35}, uncorrected NFT collapses, while
GuidanceNFT steadily improves GenEval reward
(Figure~\ref{fig:explicit-guidance-diagnosis}(b)). Replacing the center
with the analytic CFG field accelerates collapse. The corresponding
diagnostics reveal the same ordering: the analytic center develops
the largest gap to its fitted rollout reference, the uncorrected
center develops a smaller but growing gap, and GuidanceNFT maintains a
small gap throughout training (panel (c)). These results connect the
update bias in Equation~\ref{eq:center-error} to the observed
instability and show why the center must be estimated from the
re-noised rollout distribution.

The benefit extends beyond preventing collapse: GuidanceNFT combines
higher target reward with stronger performance beyond that reward
(Table~\ref{tab:sd35-cfg4-update200}). We specialize SD3.5-M for
GenEval, OCR, or Learned3, which combines PickScore, HPSv2.1, and
CLIPScore. Adding CFG after standard NFT reduces the on-domain reward
for every objective. GuidanceNFT instead improves all three on-domain
rewards over NFT both with and without CFG. For GenEval and OCR
specialization, it also outperforms both NFT variants on all seven
evaluation metrics. The OCR
model illustrates the scale of this improvement: held-out GenEval
success rises from 0.035 to 0.488, while OCR reward increases from
0.971 to 0.975. Learned3 fine-tuning further improves Aesthetic and
ImageReward, two evaluators excluded from its training objective.
Online center correction therefore brings the benefits of guidance
into reward optimization itself.

\input{tables/sd35_full_cfg_update200}

\begin{figure}[t]
\centering
\includegraphics[width=\linewidth]{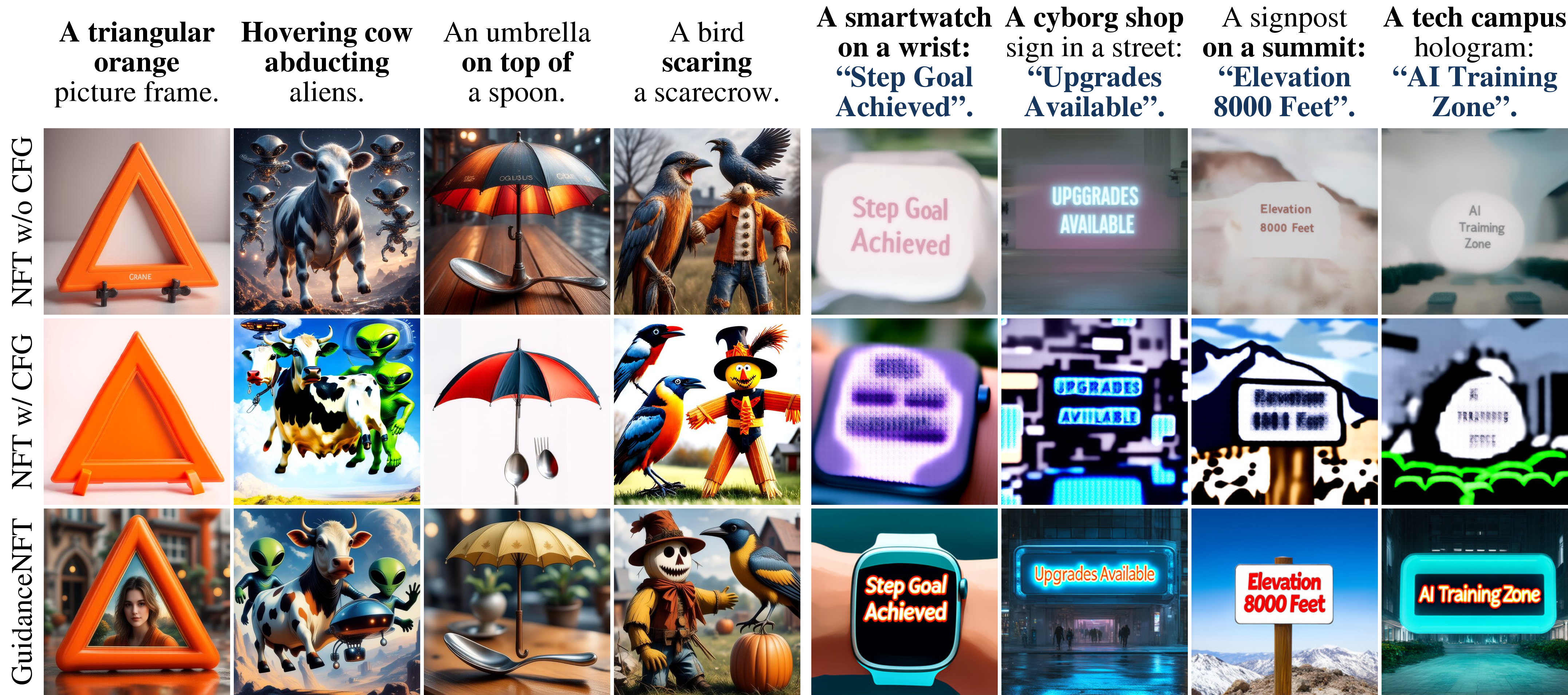}
\caption{\textbf{GuidanceNFT preserves prompt content beyond the training reward.}
Learned3 examples (first \rev{four} columns) show improved object and spatial
alignment; OCR examples (last \rev{four}) combine legible text with the
requested scene. Standard NFT loses scene content in the OCR examples,
and adding CFG damages text legibility. Rows compare NFT without CFG,
NFT with CFG 4, and GuidanceNFT with CFG 4. Full prompts and selection details
are in Appendix~\ref{app:sd35-qualitative}.}
\label{fig:sd35-learned3-ocr-qualitative}
\end{figure}

The qualitative results make this distinction visible
(Figure~\ref{fig:sd35-learned3-ocr-qualitative}). GuidanceNFT improves object
and spatial alignment in the Learned3 examples, while adding CFG to
NFT can disrupt image content. For OCR, standard NFT renders the
requested words but largely loses the surrounding scene. GuidanceNFT
combines legible text with recognizable scene context in all \rev{four}
displayed examples, including a watch on a wrist and a sign on a
mountain summit. These examples illustrate how guided fine-tuning can
preserve full-prompt fidelity alongside a specialized reward.

\subsection{Built-in Guidance: Model Guidance and Self-Conditioning}
\label{sec:intrinsic-experiments}

Guidance need not be an external sampling operation for center
mismatch to matter. We next apply the same correction to Model
Guidance (MG)~\citep{tang2025modelguidance} and ELF self-conditioning
guidance~\citep{hu2026elf}, whose native predictions already include
guidance.

\begin{figure}[t]
\centering
\includegraphics[width=\linewidth]{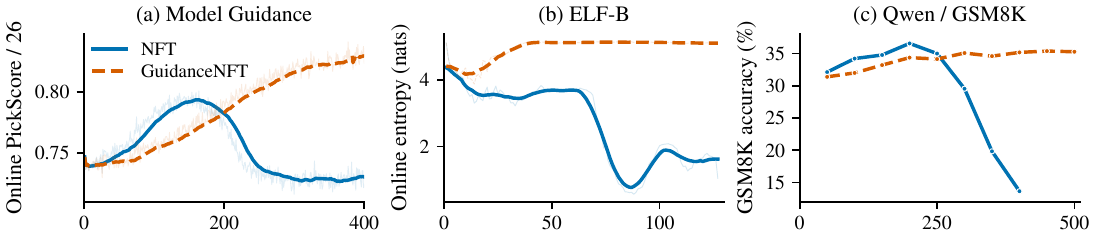}
\caption{\textbf{Center correction stabilizes NFT with built-in guidance.}
GuidanceNFT sustains training on Model Guidance, ELF-B, and GSM8K, where
standard NFT loses reward or collapses. Faint image/text traces show
raw rewards and bold curves show smoothed trends; reasoning markers
denote evaluated checkpoints. \rev{Horizontal axes show training updates.} Protocols are in
Appendix~\ref{app:experimental-protocol}.}
\label{fig:intrinsic-guidance-dynamics}
\label{fig:sccfg-training-dynamics}
\end{figure}

\begin{figure}[t]
\centering
\includegraphics[width=\linewidth]{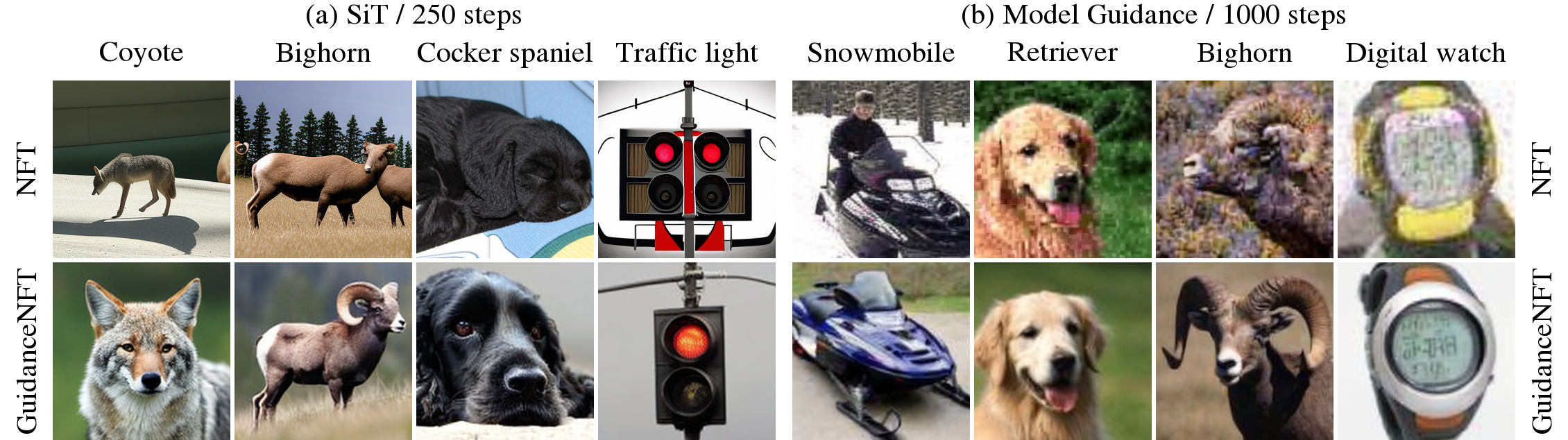}
\caption{\textbf{GuidanceNFT improves image quality beyond NFT's reward peak.}
Selected SiT (left) and Model Guidance (right) samples show more detailed,
recognizable class instances after center correction. NFT uses its
update-200 checkpoint before degradation; GuidanceNFT uses update 400.
Displayed images use 250 (SiT) and 1,000 (MG)
sampling steps; selection details are in Appendix~\ref{app:sit-checkpoints}.}
\label{fig:mg-qualitative}
\end{figure}

Across these models, center correction consistently stabilizes
training (Figure~\ref{fig:intrinsic-guidance-dynamics}). Standard NFT
loses reward on MG, degenerates into repetitive text on ELF-B, and
suffers a sharp reasoning-accuracy drop. GuidanceNFT sustains training in
all three settings. On MG, continued improvement yields a higher
reward than NFT's best retained checkpoint; the same advantage holds
for SiT (Appendix~\ref{app:sit-checkpoints}). The accompanying image
comparison shows more detailed and recognizable class instances
(Figure~\ref{fig:mg-qualitative}).
{\revon These reward gains do not come at the cost of distributional
fidelity. Because reward fine-tuning intentionally moves samples toward
higher reward, we compare each model with an ImageNet reference tilted
toward high PickScore (Appendix~\ref{app:tilted-fid}). We compare both methods
at update 200, where NFT's reward peaks before it degrades. There,
GuidanceNFT attains a higher PickScore and Inception Score and
a lower tilted FID than NFT in all three settings, and on SiT with CFG 1
its tilted FID is also lower than the pretrained model's (Table~\ref{tab:imagenet-tilted-fid}).\par}

\input{tables/imagenet_tilted_fid}

\begin{figure}[t]
\centering
\newsavebox{\elfexamplesbox}
\sbox{\elfexamplesbox}{%
\begin{minipage}[c]{0.57\linewidth}
\input{tables/sccfg_qualitative_examples}
\end{minipage}}
\usebox{\elfexamplesbox}\hfill
\begin{minipage}[c]{0.41\linewidth}
\raggedleft
\includegraphics[height=\dimexpr\ht\elfexamplesbox+\dp\elfexamplesbox\relax]{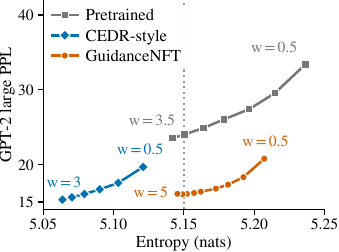}
\end{minipage}
\caption{\textbf{GuidanceNFT improves the quality--diversity tradeoff on ELF-B.}
Left: GuidanceNFT preserves varied text where standard NFT degenerates into
repetition; both train at SCCFG 3. Right: GuidanceNFT retains higher entropy
at comparable perplexity than the pretrained model and CEDR-style
reconstruction. GuidanceNFT and reconstruction train at SCCFG 3 and 1,
respectively. Markers denote evaluated guidance settings, with
$w$ values labeled; the dotted line marks entropy 5.15.
}
\label{fig:sccfg-main-sweep}
\label{fig:sccfg-qualitative-collapse}
\end{figure}

On ELF-B, stable training translates into a better balance between
language quality and diversity (Figure~\ref{fig:sccfg-main-sweep}).
Fine-tuning on OpenWebText with GPT-2-large log-likelihood reward causes
standard NFT to repeat short phrases, whereas GuidanceNFT retains varied
text. We further compare with the guidance-aware velocity
reconstruction from CEDR~\citep{anonymous2026lfrm}, adapted from
self-conditioned reasoning to ELF-B. Across evaluation guidance
settings, GuidanceNFT achieves a better perplexity--entropy tradeoff than
both this baseline and the pretrained model, retaining higher entropy
at comparable perplexity. Center correction thus enables reward
optimization without the repetition collapse of standard NFT.

The reasoning results show that center correction improves task
performance as well as training stability. On the Qwen-based ELF
model, GuidanceNFT outperforms both NFT's best retained checkpoint and the
CEDR reasoning checkpoint at every reported Pass@$k$ on
GSM8K~\citep{cobbe2021gsm8k}
(Table~\ref{tab:qwen-full1319-nft-comparison}). The gains over NFT are
1.35 percentage points in Pass@1 and 1.59 in Pass@8. Under this
evaluation protocol, the benefit therefore extends beyond avoiding
late collapse: GuidanceNFT exceeds the performance of NFT before its
decline.

\input{tables/qwen_full1319_nft_comparison}

%% file: tables/sd35_full_cfg_update200.tex
\begin{table}[!b]
\centering
\small
\caption{\textbf{SD3.5-M reward improvement and quality retention.} W/o and w/ CFG denote evaluation scales 1 and 4. Fine-tuned rows average evaluation seeds 42 and 123; pretrained rows use seed 42. Train reward measures the optimized objective on its test set. Gray marks training criteria; bold marks the best within each objective. Seed stds are given in Appendix~\ref{app:sd35-tables}.}
\label{tab:sd35-cfg4-update200}
\label{tab:sd35-full-cfg-update200}
\setlength{\tabcolsep}{2pt}
\renewcommand{\arraystretch}{1.06}
\edef\sdrestorerulesep{\global\aboverulesep=\the\aboverulesep\global\belowrulesep=\the\belowrulesep\relax}
\edef\sdtop{\noexpand\rule{0pt}{\noexpand\dimexpr\arraystretch\noexpand\ht\noexpand\strutbox+\the\belowrulesep\relax}}
\edef\sdbot{\noexpand\rule[-\noexpand\dimexpr\arraystretch\noexpand\dp\noexpand\strutbox+\the\aboverulesep\relax]{0pt}{0pt}}
\begin{tabular}{@{}p{36pt}p{66pt}>{\centering\arraybackslash}p{33pt}*{7}{>{\centering\arraybackslash}p{\dimexpr(\linewidth-135pt-18\tabcolsep)/7\relax}}@{}}
\toprule
\multirow{2}{*}{\shortstack[l]{\textbf{Training}\\\textbf{objective}}} & \multirow{2}{*}{\textbf{Method}} & \multirow{2}{*}{\shortstack{\textbf{Train}\\\textbf{reward}}} & \multicolumn{7}{c}{\textbf{Held-out evaluation}} \\
\cmidrule(l){4-10}
 & & & \makebox[\linewidth][c]{{\fontsize{7.2}{9}\selectfont\bfseries GenEval}} & \makebox[\linewidth][c]{{\fontsize{7.2}{9}\selectfont\bfseries OCR}} & \makebox[\linewidth][c]{{\fontsize{7.2}{9}\selectfont\bfseries PickScore}} & \makebox[\linewidth][c]{{\fontsize{7.2}{9}\selectfont\bfseries CLIPScore}} & \makebox[\linewidth][c]{{\fontsize{7.2}{9}\selectfont\bfseries HPSv2.1}} & \makebox[\linewidth][c]{{\fontsize{7.2}{9}\selectfont\bfseries Aesthetic}} & \makebox[\linewidth][c]{{\fontsize{7.2}{9}\selectfont\bfseries ImgRwd}} \\
\midrule
\noalign{\global\aboverulesep=0pt\global\belowrulesep=0pt}
\multirow{2}{*}{---} & Pretrained {\scriptsize w/o CFG} & --- & 0.283 & 0.113 & 20.69 & 0.232 & 0.199 & 4.990 & -0.594 \\
 & \sdbot Pretrained {\scriptsize w/ CFG} & --- & 0.636 & 0.532 & 22.38 & 0.278 & 0.276 & 5.252 & 0.820 \\
\midrule
\multirow{3}{*}{GenEval} & \sdtop NFT {\scriptsize w/o CFG} & 0.900 & \cellcolor{gray!15}0.855 & 0.434 & 21.70 & 0.265 & 0.234 & 4.886 & 0.271 \\
 & NFT {\scriptsize w/ CFG} & 0.657 & \cellcolor{gray!15}0.565 & 0.624 & 21.19 & 0.271 & 0.216 & 4.773 & 0.148 \\
 & \sdbot GuidanceNFT & \textbf{0.914} & \cellcolor{gray!15}\textbf{0.877} & \textbf{0.674} & \textbf{21.83} & \textbf{0.277} & \textbf{0.250} & \textbf{4.947} & \textbf{0.682} \\
\midrule
\multirow{3}{*}{OCR} & \sdtop NFT {\scriptsize w/o CFG} & 0.971 & 0.035 & \cellcolor{gray!15}0.971 & 19.61 & 0.208 & 0.137 & 4.208 & -1.679 \\
 & NFT {\scriptsize w/ CFG} & 0.192 & 0.265 & \cellcolor{gray!15}0.235 & 20.21 & 0.242 & 0.188 & 4.596 & -0.919 \\
 & \sdbot GuidanceNFT & \textbf{0.975} & \textbf{0.488} & \cellcolor{gray!15}\textbf{0.974} & \textbf{21.83} & \textbf{0.274} & \textbf{0.256} & \textbf{5.120} & \textbf{0.425} \\
\midrule
\multirow{3}{*}{Learned3} & \sdtop NFT {\scriptsize w/o CFG} & 1.519 & 0.671 & 0.497 & \cellcolor{gray!15}\textbf{23.51} & \cellcolor{gray!15}0.284 & \cellcolor{gray!15}\textbf{0.332} & 5.606 & 1.397 \\
 & NFT {\scriptsize w/ CFG} & 1.495 & \textbf{0.701} & 0.579 & \cellcolor{gray!15}22.99 & \cellcolor{gray!15}0.290 & \cellcolor{gray!15}0.326 & 5.397 & 1.414 \\
 & \sdbot GuidanceNFT & \textbf{1.528} & 0.630 & \textbf{0.692} & \cellcolor{gray!15}23.24 & \cellcolor{gray!15}\textbf{0.292} & \cellcolor{gray!15}\textbf{0.332} & \textbf{5.653} & \textbf{1.448} \\
\bottomrule
\end{tabular}
\sdrestorerulesep
\end{table}

%% file: tables/imagenet_tilted_fid.tex
\begin{table}[b]
\centering
\small
\setlength{\tabcolsep}{4pt}
\caption{{\revon\textbf{ImageNet generation quality at update 200.}
PickScore, Inception Score (IS), and reward-tilted FID (tFID) of 10,000
250-step samples; tFID compares samples with a PickScore-tilted ImageNet
reference (Appendix~\ref{app:tilted-fid}). Fine-tuned models use update
200, where NFT's reward peaks. Bold marks the better fine-tuned method.}}
\label{tab:imagenet-tilted-fid}
\begin{tabular}{@{}lccccccccc@{}}
\toprule
& \multicolumn{3}{c}{SiT, CFG 1} & \multicolumn{3}{c}{SiT, CFG 1.5} & \multicolumn{3}{c}{Model Guidance} \\
\cmidrule(lr){2-4}\cmidrule(lr){5-7}\cmidrule(l){8-10}
Method & PickScore$\uparrow$ & IS$\uparrow$ & tFID$\downarrow$
 & PickScore$\uparrow$ & IS$\uparrow$ & tFID$\downarrow$
 & PickScore$\uparrow$ & IS$\uparrow$ & tFID$\downarrow$ \\
\midrule
Pretrained & 19.885 & 115.3 & 11.28 & 20.321 & 218.7 & 4.28 & 20.338 & 240.6 & 3.93 \\
\midrule
NFT & 20.398 & 111.0 & 17.64 & 20.645 & 181.9 & 11.58 & 20.335 & 94.8 & 30.61 \\
GuidanceNFT & \textbf{20.907} & \textbf{200.4} & \textbf{7.73} & \textbf{21.113} & \textbf{270.1} & \textbf{6.76} & \textbf{21.017} & \textbf{231.4} & \textbf{13.30} \\
\bottomrule
\end{tabular}
\end{table}

%% file: tables/sccfg_qualitative_examples.tex
\begingroup
\setlength{\fboxsep}{3pt}
\noindent\fcolorbox{black!25}{blue!6}{%
\parbox{\dimexpr\linewidth-2\fboxsep-2\fboxrule\relax}{%
\raggedright\small
\textbf{NFT}\hfill {\scriptsize Entropy: 2.06\quad Gen.\ PPL: 1.56}\\[2pt]
{\rmfamily health care and health care for children’s care for children’s health care for children’s health care for children’ss health care for children’s health care for children’s health care and health care for children’ss health care for children’s~\ldots\strut}
}}
\par\vspace{4pt}
\noindent\fcolorbox{black!25}{orange!10}{%
\parbox{\dimexpr\linewidth-2\fboxsep-2\fboxrule\relax}{%
\raggedright\small
\textbf{GuidanceNFT}\hfill {\scriptsize Entropy: 5.14\quad Gen.\ PPL: 16.40}\\[2pt]
{\rmfamily FireAvener is a non-profit organization created to promote magazine, film, radio and television shows around the world. Their shows reach a range of audiences around the world and focused on poverty, poverty and suffering.~\ldots\strut}
}}
\par
\endgroup

%% file: tables/qwen_full1319_nft_comparison.tex
\begin{table}[t]
\centering\small
\caption{\textbf{GSM8K with SCCFG-2 evaluation.} Pass@$k$ (\%): at least one correct answer in $k$ attempts. Selected checkpoints are within 500 updates, including NFT's best retained point. Values are mean $\pm$ problem-bootstrap std (Appendix~\ref{app:qwen-sccfg-online}).}
\label{tab:qwen-full1319-nft-comparison}
\label{tab:qwen-intermediate-k}
\renewcommand{\arraystretch}{1.08}
\setlength{\tabcolsep}{5pt}
\begin{tabular}{@{}lcccc@{}}
\toprule
Method & Pass@1 & Pass@2 & Pass@4 & Pass@8 \\
\midrule
Pretrained & 39.17 {\footnotesize $\pm$ 1.04} & 49.82 {\footnotesize $\pm$ 1.15} & 59.20 {\footnotesize $\pm$ 1.21} & 67.32 {\footnotesize $\pm$ 1.29} \\
NFT & 41.68 {\footnotesize $\pm$ 1.08} & 51.53 {\footnotesize $\pm$ 1.17} & 59.78 {\footnotesize $\pm$ 1.22} & 67.02 {\footnotesize $\pm$ 1.30} \\
CEDR & 42.16 {\footnotesize $\pm$ 1.11} & 51.36 {\footnotesize $\pm$ 1.19} & 59.40 {\footnotesize $\pm$ 1.24} & 66.26 {\footnotesize $\pm$ 1.32} \\
GuidanceNFT & \textbf{43.03} {\footnotesize $\pm$ 1.10} & \textbf{52.66} {\footnotesize $\pm$ 1.16} & \textbf{61.24} {\footnotesize $\pm$ 1.20} & \textbf{68.61} {\footnotesize $\pm$ 1.26} \\
\bottomrule
\end{tabular}
\end{table}

%% file: related_work.tex
\section{Related Work}
\label{sec:related-work}

\paragraph{Regression-based reward fine-tuning.}
Forward regression offers an alternative to the trajectory policy
gradients used by DDPO~\citep{black2024ddpo} and
Flow-GRPO~\citep{liu2025flowgrpo}. DiffusionNFT~\citep{zheng2026diffusionnft}
learns from positive and negative reward-weighted populations, while
Reward Score Matching~\citep{lee2026rsm},
AdvantageFlow~\citep{kveton2026advantageflow}, Reinforce Adjoint
Matching~\citep{bergmeister2026ram}, and reward-based velocity matching
(RVM)~\citep{choi2026rvm} develop related regression formulations.
RVM recovers NFT through reward weights and anchor choices. We
identify the center required by guided NFT and show how its error
distorts the reward-driven update.

\paragraph{Guidance and the generated distribution.}
CFG~\citep{ho2022cfg} combines conditional and unconditional
predictions, but the resulting sampler need not follow the density
suggested by the combined
scores~\citep{bradley2025predictor,chidambaram2024guidance,jiang2026analyticcfg}.
RVM likewise notes that a learned field can differ from the mean
velocity induced by its endpoints~\citep{choi2026rvm}. We establish
the consequence for NFT and correct its center online, including
when guidance is built into the
model~\citep{tang2025modelguidance,hu2026elf}.

\paragraph{Estimating the generated distribution.}
Auxiliary scores of generated samples appear in variational score
distillation~\citep{wang2023prolificdreamer},
distribution matching distillation (DMD)~\citep{yin2024dmd}, and
DMD2~\citep{yin2024dmd2}, which connects estimator error to training
instability. We use flow matching~\citep{lipman2023flowmatching} to
estimate NFT's center. Unlike guided teacher
distillation~\citep{meng2023guided} or trajectory straightening in
rectified flow~\citep{liu2023rectified}, this auxiliary fit calibrates
the reward objective and is discarded at inference.

\paragraph{Complementary adaptations and failure mechanisms.}
Centered Reward Distillation~\citep{zhu2026crd} controls drift with
within-prompt reward centering and a guided pretrained reference;
its reward centering differs from our regression center.
Flow-OPD~\citep{fang2026flowopd}
addresses multi-reward interference by distilling specialized
teachers. ReNFT~\citep{bao2026renft} repairs diversity collapse after
reward fine-tuning by generating counterfactual samples through
internal model routes; our correction addresses center-induced bias
during guided NFT training. Pretrained unconditional
predictions~\citep{phunyaphibarn2026unconditional} can help retain
generation quality, while self-conditioning
feedback~\citep{zhang2026repetition} and training--sampling
mismatch~\citep{li2026untied} affect language flows. Our experiments
include a pretrained-empty control and the SCCFG-compatible NFT
adaptation of \citet{anonymous2026lfrm}, denoted CEDR.

%% file: conclusion.tex
\section{Conclusion}

The accuracy of NFT's regression center is critical to learning from
guided samples. A center gap introduces an additional component into
the reward-driven update, explaining why matching sampling and
training guidance alone is insufficient. GuidanceNFT addresses this gap
with online flow matching and a shared correction to NFT's positive
and negative regressions. Across explicit and built-in guidance,
the correction prevents the training degradation observed with NFT
and improves reward optimization, image quality, and language
generation. By separating center estimation from the policy update,
GuidanceNFT makes guided reward fine-tuning effective without additional
inference passes.
\label{sec:conclusion-end}

%% file: statements.tex
\section*{Reproducibility Statement}
The appendices specify the population assumptions, training and evaluation
protocols, checkpoint-selection rules, and uncertainty calculations.
Analysis scripts trace the reported summaries to recorded evaluations
and distinguish training runs from evaluation seeds.

\section*{AI Use Statement}
Generative AI tools assisted with literature review, manuscript structure
and editing, mathematical checks, experiment design, analysis code,
statistical summaries, and figure preparation. The reported model results
are drawn from recorded evaluations; analytical diagnostics are identified
separately. This statement describes the assistance used in preparing the
present revision.

%% file: appendix.tex
\section{Derivations for GuidanceNFT}
\label{app:nft-theory}

We derive the center required by NFT, quantify the bias caused by a center
gap, and connect center estimation to the resulting policy update.
The rollout law and reward weights are fixed within each update; all
second moments are finite. Population optima are taken over unrestricted
square-integrable fields, with $\beta>0$. For Gaussian noising, assume differentiable
schedules and sufficient regularity to differentiate under the integral:
\begin{equation}
X_t=a_tX+\sigma_t\epsilon,\qquad
U=\dot a_tX+\dot\sigma_t\epsilon,\qquad
\epsilon\sim\mathcal N(0,I).
\label{eq:training-forward-process}
\end{equation}
Here $c\sim p(c)$, $X\sim q(\cdot\mid c)$, and $t\sim p(t)$, with
$t$ and $\epsilon$ independent of $X$. Time runs from noise to data,
and $a_t,\sigma_t>0$ at interior times. We write $z=(X_t,c,t)$,
$q=q_{\rm CFG}$, $C_q=\bar v_{q_{\rm CFG}}$, $C_\phi=v_\phi$, and
$d=v_\theta-\operatorname{sg}(v^{\rm old})$.
The target $U$ is the sample velocity $v$ in the main text.

\subsection{Reward-Weighted Targets and the Center Error}

Let $Z(c)=\mathbb E_q[r\mid c]\in(0,1)$ and define
\begin{equation}
q^+(X\mid c)=\frac{r(X,c)q(X\mid c)}{Z(c)},\qquad
q^-(X\mid c)=\frac{[1-r(X,c)]q(X\mid c)}{1-Z(c)}.
\label{eq:reward-endpoint-distributions}
\end{equation}
At a noisy state, their regression targets are
\begin{equation}
\alpha(z)=\mathbb E[r\mid z],\qquad
v^+(z)=\frac{\mathbb E[rU\mid z]}{\alpha(z)},\qquad
v^-(z)=\frac{\mathbb E[(1-r)U\mid z]}{1-\alpha(z)}.
\label{eq:positive-negative-fields}
\end{equation}
Positive Gaussian kernels ensure $0<\alpha<1$. Weighted least squares
identifies these fields as the two optimal predictions because
\begin{equation}
\mathbb E[r\|h-U\|^2\mid z]
=\alpha\|h-v^+\|^2+\mathbb E[r\|U-v^+\|^2\mid z],
\label{eq:weighted-regression}
\end{equation}
and the analogous identity holds for $1-r$. The cross term vanishes
by $\mathbb E[r(U-v^+)\mid z]=0$.
Adding the two weighted conditional means gives
\begin{equation}
C_q=\mathbb E[U\mid z]=\alpha v^++(1-\alpha)v^-,\qquad
\Delta=v^+-v^-.
\label{eq:appendix-center-balance}
\end{equation}
The center is uniquely determined by the weighted branch balance:
\begin{equation}
\alpha(v^+-C)=(1-\alpha)(C-v^-)
\quad\Longleftrightarrow\quad
C=\alpha v^++(1-\alpha)v^-=C_q.
\label{eq:unique-center-balance}
\end{equation}
At this center, both sides equal $\alpha(1-\alpha)\Delta$.
Thus $v^+-C_q=(1-\alpha)\Delta$ and $C_q-v^-=\alpha\Delta$:
the unique matched center calibrates both branches to the same
reward-improvement direction with the correct population weights.

For an arbitrary frozen center $C$, NFT minimizes
\begin{equation}
\mathcal L(C;d)=\mathbb E\!\left[
r\|C+\beta d-U\|^2+(1-r)\|C-\beta d-U\|^2\right].
\label{eq:shifted-nft-loss}
\end{equation}
The shared correction in Equation~\ref{eq:corrected-nft-objective}
sets $C=v^{\rm old}+\mathcal C$. Applying the weighted-regression
identity reduces its conditional objective, up to terms independent of
$d$, to
\begin{equation}
\ell_z(C;d)=\alpha\|C+\beta d-v^+\|^2
 +(1-\alpha)\|C-\beta d-v^-\|^2.
\label{eq:nft-population-targets}
\end{equation}
Its gradient and unique minimizer satisfy
\begin{equation}
\begin{aligned}
\nabla_d\ell_z
 &=2\beta\big[\beta d+(2\alpha-1)C-\alpha v^++(1-\alpha)v^-\big],\\
\beta d_C^*
 &=\alpha(v^+-C)+(1-\alpha)(C-v^-)\\
 &=2\alpha(1-\alpha)\Delta+(1-2\alpha)(C-C_q).
\end{aligned}
\label{eq:appendix-center-error}
\end{equation}
This proves Equation~\ref{eq:center-error}. Center error contributes a
separate displacement, rather than simply rescaling the intended one.
At states with $\alpha\ne1/2$, exact recovery of the intended displacement
requires $C=C_q$; at $\alpha=1/2$, the center term cancels.
The latter cancellation concerns their sum; the weighted branch-balance
condition still selects $C_q$ uniquely.

For $C=C_q$, completing the square gives
\begin{equation}
\mathcal L(C_q;d)
 =\beta^2\mathbb E\left\|d-\frac{2\alpha(1-\alpha)}{\beta}\Delta\right\|^2
 +\overline K,
\label{eq:nft-displacement-regression}
\end{equation}
where $\overline K$ is independent of $d$. Consequently,
\begin{equation}
v_{\theta^*}=v^{\rm old}+\frac{2\alpha(1-\alpha)}{\beta}\Delta.
\label{eq:matched-velocity-update}
\end{equation}
The correction changes the regression center while retaining the old
policy as the baseline of the update.

A useful equivalent form uses $A=2r-1$:
\begin{equation}
\begin{aligned}
r\|C+\beta d-U\|^2+(1-r)\|C-\beta d-U\|^2
 &=\|C-U\|^2+\beta^2\|d\|^2+2\beta A\langle C-U,d\rangle\\
 &=\|\beta d-A(U-C)\|^2+(1-A^2)\|U-C\|^2.
\end{aligned}
\label{eq:advantage-expansion-proof}
\end{equation}
The last term is constant in $d$, but changes with $C$. This form and
the covariance identity below also apply when $Z=0$ or $1$, without
introducing an undefined tilted population.

\subsection{The Score Represented by the Center}

Bayes' rule gives the local positive weight
\begin{equation}
\alpha(z)=\frac{Z(c)q_t^+(x\mid c)}{q_t(x\mid c)},\qquad
1-\alpha(z)=\frac{[1-Z(c)]q_t^-(x\mid c)}{q_t(x\mid c)}.
\label{eq:posterior-reward-weight}
\end{equation}
For $K_t(x\mid X)=\mathcal N(x;a_tX,\sigma_t^2I)$,
differentiating $q_t(x\mid c)=\int K_t(x\mid X)q(X\mid c)\,dX$ yields
\begin{equation}
\nabla_x\log q_t(x\mid c)
=-\frac{x-a_t\mathbb E[X\mid z]}{\sigma_t^2}
=-\frac{\mathbb E[\epsilon\mid z]}{\sigma_t}.
\end{equation}
Substitution into $C_q=\dot a_t\mathbb E[X\mid z]
+\dot\sigma_t\mathbb E[\epsilon\mid z]$ gives
\begin{equation}
C_q(x,t,c)=A_tx+B_t\nabla_x\log q_t(x\mid c),\qquad
A_t=\frac{\dot a_t}{a_t},\quad
B_t=\frac{\dot a_t\sigma_t^2}{a_t}-\sigma_t\dot\sigma_t.
\label{eq:center-score}
\end{equation}
The velocity--score correspondence is invertible when $B_t\ne0$.
Applying it to $q^\pm$ gives
\begin{equation}
v^\pm=A_tx+B_t\nabla_x\log q_t^\pm,\qquad
\Delta=B_t\nabla_x\log\frac{q_t^+}{q_t^-}.
\label{eq:positive-negative-scores}
\end{equation}
Using $q_t^+=\alpha q_t/Z$ then relates the weighted direction to the
posterior reward:
\begin{equation}
\alpha(1-\alpha)\Delta
 =B_t\nabla_x\alpha
 =\operatorname{Cov}(r,U\mid z).
\label{eq:reward-direction-score}
\end{equation}
Centering therefore removes the reward-independent mean velocity and
extracts a reward-correlated change. For $a_t=t$, $\sigma_t=1-t$,
$B_t=(1-t)/t>0$, so the intended displacement points toward increasing
posterior reward. These identities concern the forward-noised endpoint
law, regardless of the sampler trajectories that generated its samples.

\subsection{Why a Guided Score Need Not Match Its Rollout Law}
\label{app:guided-score-mismatch}

For exact component scores $s_g=\nabla\log p_{g,t}$ and
$s_b=\nabla\log p_{b,t}$, CFG gives
\begin{equation}
w s_g+(1-w)s_b
=\nabla\log\!\left(p_{g,t}^{\,w}p_{b,t}^{\,1-w}\right).
\label{eq:cfg-product-score}
\end{equation}
A score at each time need not define a family consistent with a single
endpoint law under forward noising~\citep{bradley2025predictor, chidambaram2024guidance,
jiang2026analyticcfg}. To see the distinction directly,
consider the special case of additive Gaussian noising, parameterized by
increasing noise variance $\tau$. Both component densities satisfy the
heat equation $\partial_\tau p=\tfrac12\nabla^2p$. For
$h_\tau=p_{g,\tau}^{\,w}p_{b,\tau}^{\,1-w}$, direct differentiation gives
\begin{equation}
\frac{\partial_\tau h_\tau-\tfrac12\nabla^2h_\tau}{h_\tau}
=\frac{w(1-w)}{2}\|s_g-s_b\|^2.
\label{eq:cfg-forward-inconsistency}
\end{equation}
When the right-hand side varies with $x$, a time-dependent normalizer
cannot remove it: even a normalizable guided density need not solve the
same forward equation. Special component families can satisfy it.

\paragraph{An exact Gaussian mismatch.}
Take $a_t=t$, $\sigma_t=1-t$, component endpoint laws
$\mathcal N(0,1)$ and $\mathcal N(0,4)$, and CFG scale $w=2$.
Write $D_g=t^2+(1-t)^2$ and $D_b=4t^2+(1-t)^2$.
The component velocities are
$v_g=\tfrac12\partial_t\log D_g\,x$ and
$v_b=\tfrac12\partial_t\log D_b\,x$.
Starting from $\mathcal N(0,1)$, the guided ODE with
$g=2v_g-v_b$ therefore has variance $D_g^2/D_b$, and its
endpoint law is $q=\mathcal N(0,1/4)$.
Re-noising that endpoint law instead gives variance
$t^2/4+(1-t)^2$. At $t=1/4$,
\begin{equation}
\begin{aligned}
\operatorname{Var}(X_t^{\rm sampler})&=25/52,
&\operatorname{Var}(X_t^{\rm renoise})&=37/64,\\
g(x,t)&=-\frac{124}{65}x,
&C_q(x,t)&=-\frac{44}{37}x,\\
e(x,t)&=g(x,t)-C_q(x,t)=-\frac{1728}{2405}x.
\end{aligned}
\label{eq:gaussian-anchor-mismatch}
\end{equation}
No neural approximation or numerical solver is involved. These are
also compatible conditional and unconditional laws: assign the first component class probability
$1/4$ and use $[\mathcal N(0,4)-(1/4)\mathcal N(0,1)]/(3/4)$
for the other class, which is a nonnegative normalized density.
The example instantiates the Gaussian guidance analyses cited above;
its implication for NFT is the additional term in
Equation~\ref{eq:center-error}.

\subsection{Center Accuracy and Parameter Updates}
\label{app:center-optimization}

For an erroneous center $C=C_q+e$, insert
$d_C^*-d_{C_q}^*=(1-2\alpha)e/\beta$ into the matched-center
quadratic. Subtracting its minimum and averaging over $z$ gives
\begin{equation}
\begin{aligned}
\mathcal L(C_q;d_C^*)-\mathcal L(C_q;d_{C_q}^*)
 &=\beta^2\mathbb E\|d_C^*-d_{C_q}^*\|^2\\
 &=\mathbb E[(1-2\alpha)^2\|C-C_q\|^2].
\end{aligned}
\label{eq:anchor-weighted-excess-risk}
\end{equation}
Both losses on the left use $C_q$; comparing losses with different
centers would also change their additive constants. The $1/\beta$
displacement bias cancels the quadratic's $\beta^2$ scaling.

The flow-matching objective in Equation~\ref{eq:center-fm} has the
standard conditional-mean decomposition~\citep{lipman2023flowmatching}:
\begin{equation}
\mathcal L_{\rm FM}(\phi)
=\mathbb E\|C_\phi-C_q\|^2+\mathbb E\|U-C_q\|^2.
\label{eq:center-fm-decomposition}
\end{equation}
For two frozen predictors evaluated under the same law, writing
$R_q(h)=\mathbb E\|h(z)-U\|^2$ gives
\begin{equation}
R_q(h_1)-R_q(h_2)
=\mathbb E\|h_1-C_q\|^2-\mathbb E\|h_2-C_q\|^2.
\label{eq:paired-fm-risk}
\end{equation}
An empirical comparison uses the same held-out endpoints, times,
noise, masks, and conditioning inputs for both predictors. A positive
risk difference supports smaller center MSE for $h_2$ without estimating
the irreducible variance.

For $e=C_\phi-C_q$, Equation~\ref{eq:center-error} gives
\begin{equation}
\mathbb E\|d_{C_\phi}^*-d_{C_q}^*\|^2
=\frac{1}{\beta^2}\mathbb E[(1-2\alpha)^2\|e\|^2]
\leq\frac{R_q(C_\phi)-R_q(C_q)}{\beta^2}.
\label{eq:center-risk-bound}
\end{equation}
FM excess risk bounds the target error.

The same decomposition separates center approximation from finite
policy fitting. Let
$\epsilon_C=R_q(C_\phi)-R_q(C_q)$ and
$\epsilon_N=\mathcal L(C_\phi;\widehat d)-\inf_d\mathcal L(C_\phi;d)$,
where the infimum is over unrestricted square-integrable fields.
Completing the conditional quadratic gives
\begin{equation}
\mathcal L(C;d)=\beta^2\mathbb E\|d-d_C^*\|^2+K_C,
\qquad
\left\|\widehat d-\frac{2\alpha(1-\alpha)\Delta}{\beta}\right\|_{L^2}
\leq\frac{\sqrt{\epsilon_C}+\sqrt{\epsilon_N}}{\beta}.
\label{eq:finite-fit-displacement-bound}
\end{equation}
Here $K_C$ does not depend on $d$, and the norm uses the re-noised
training law. The bound follows from the triangle inequality and
Equation~\ref{eq:center-risk-bound}. The policy excess risk
$\epsilon_N$ includes representation and optimization error.

For a differentiable recipient with
$J_\theta(z)=\partial v_\theta(z)/\partial\theta$, differentiate
Equation~\ref{eq:advantage-expansion-proof} while holding the rollout
law, weights, center, and baseline fixed. At the same policy, the
erroneous- and matched-center gradients differ by
\begin{equation}
\nabla_\theta\mathcal L(C;d)-\nabla_\theta\mathcal L(C_q;d)
=2\beta\mathbb E[(2\alpha-1)J_\theta^\top e].
\label{eq:center-parameter-gradient}
\end{equation}
This describes the current gradient, without assuming convergence to
an unrestricted optimum.

The recipient field determines how this displacement changes sampling.
For native guided predictions, it is the guided-field update itself.
For conditional updates under external CFG,
\begin{equation}
\Delta v_w=w\Delta v_c+(1-w)\Delta v_\varnothing.
\label{eq:guided-field-update}
\end{equation}
The second term vanishes only for a functionally frozen empty branch.
These fixed-population identities establish update bias.

\subsection{The Training Population and Conditioning Protocol}
\label{app:training-measure}

The center and NFT objective must refer to the same base population
before the positive and negative reward weights are applied.
If endpoint filtering, replay sampling, or inserted reference solutions
change that population, write its law as $q_{\rm train}$. The preceding
identities then concern $C_{q_{\rm train}}$. A center fitted only to
unfiltered generated endpoints estimates a different conditional mean
unless these laws coincide. Match the base training measure while leaving
the center objective unweighted by reward.

Group-normalized reward weights can be represented by a frozen coefficient
$R\in[0,1]$ that depends on the endpoint and the other samples in its
group. If the re-noising variables are drawn independently after those
weights are fixed, use $r(X,c)=\mathbb E[R\mid X,c]$ in the population
endpoint laws. The conditional formulas then use
$\alpha(z)=\mathbb E[R\mid z]$. Zero mean advantage before clipping
need not persist after clipping;
even zero mean processed advantage within each group does not imply
$\alpha(z)=1/2$ at every noisy state.

For a scalar nonnegative reduction weight $m$, the corresponding
unweighted regression center is
\begin{equation}
C_m(z)=\frac{\mathbb E[mU\mid z]}{\mathbb E[m\mid z]},
\qquad \mathbb E[m\mid z]>0.
\label{eq:masked-center}
\end{equation}
Tokenwise masks give this expression coordinate by coordinate.
A positive weight determined by $z$ cancels; a mask depending on the
clean endpoint can change the center. Likewise, a deterministic self-conditioning feature computed from a
frozen model and $z$ can be treated as part of the input; stochastic
or recurrent conditioning requires its own joint law. The scalar-time
score derivation applies to the specified Gaussian kernel. Vector
clocks require the corresponding blockwise kernel and velocity
conversion. Experiments retain their documented training clocks and
conditioning protocols.

\subsection{Algebraic Centers for Intrinsic Guidance}
\label{app:intrinsic-center}

\paragraph{Model Guidance.}
Model Guidance~\citep{tang2025modelguidance} trains a conditional
prediction $M_c$ using an additive target
$U+\lambda\operatorname{sg}(M_c-M_\varnothing)$.
For a fixed coefficient $0\leq\lambda<1$, the ideal pretraining
fixed point satisfies
\begin{equation}
C_{\rm data}=(1-\lambda)M_c+\lambda M_\varnothing,\qquad
M_c=w_{\rm eff}C_{\rm data}+(1-w_{\rm eff})M_\varnothing,
\quad w_{\rm eff}=\frac{1}{1-\lambda}.
\label{eq:model-guidance-center}
\end{equation}
This reconstructs the ideal pretraining mean $C_{\rm data}$; subsequent
reward-fine-tuned rollouts generally require a different center.

\paragraph{ELF self-conditioning guidance.}
ELF~\citep{hu2026elf} uses the analogous coefficient $1-1/w$ with
self-conditioning as the guidance signal. If $S_\theta^w$ denotes
the native prediction with self-conditioning and
$S_{\theta,\varnothing}^w$ the same network evaluated with zero
self-conditioning input, the numerical reconstruction is
\begin{equation}
B_\theta^w=\frac{S_\theta^w}{w}
 +\left(1-\frac1w\right)S_{\theta,\varnothing}^w,
\qquad S_\theta^w=wB_\theta^w+(1-w)S_{\theta,\varnothing}^w.
\label{eq:sccfg-guidance-relation}
\end{equation}
This defines a reconstructed base prediction, not necessarily the output
at $w=1$ or the mean velocity of subsequent guided rollouts.

\subsection{Training Procedure}
\label{app:training-procedure}

Algorithm~\ref{alg:guided-nft} describes one outer iteration of GuidanceNFT.
The rollout law, reward coefficients, and old-policy predictions are
held fixed within the policy update. Endpoint selection, masks, and
conditioning inputs must follow Appendix~\ref{app:training-measure}
for the center objective to estimate the required center.

\begin{algorithm}[!htbp]
\caption{GuidanceNFT (one outer iteration)}
\label{alg:guided-nft}
\small
\begin{tabular}{@{}r p{0.91\linewidth}@{}}
1. & Freeze the rollout policy and baseline prediction $v^{\rm old}$.\\
2. & Generate endpoints $X$ with the specified guidance and sampler; evaluate and freeze NFT weights $r$.\\
3. & Draw fresh $(t,\epsilon)$ and construct $(X_t,U)$ by Equation~\ref{eq:training-forward-process}.\\
4. & Refresh $v_\phi$ with unweighted FM on the endpoint population used by the NFT objective.\\
5. & Freeze $v_\phi$ and form NFT's predictions $v_\theta^\pm$ from Equation~\ref{eq:nft-implicit-predictions}.\\
6. & Form $\mathcal C_\phi=\operatorname{sg}(v_\phi-v^{\rm old})$ and add it to both original predictions, $v_\theta^\pm+\mathcal C_\phi$.\\
7. & Update the policy using the NFT regressions and the stated reference penalty; refresh the rollout policy according to its schedule.\\
\end{tabular}
\end{algorithm}

\section{Toy Guidance-Aware NFT Illustration}
\label{app:toy-guidance-nft}
The tree geometry follows the Autoguidance toy construction~\citep{karras2024autoguidance}.

{\revon Figure~\ref{fig:toy-guidance-nft} compares the pretrained
model with and without CFG, NFT, guidance applied after NFT, and NFT trained
on the guided field; Table~\ref{tab:toy-guidance-nft} reports their metrics.}
All five sampled configurations use the same 2,048 evaluation noises.
Post-training guidance yields reward 0.985 and on-tree rate 71.5\%.
Guided-field training improves these to 0.996 and 90.7\%, respectively;
its on-tree rate remains below pretrained CFG's 94.6\%.

The conditional model is the class-A width-64 network; its guide is the
pretrained class-AB width-32 network. The pretrained pair remains fixed in
the CFG reference. \rev{Panel (b) applies CFG with this pretrained pair; all
guided panels use $w=3$. Panel (d)} changes only the conditional model to its ten-update
Standard NFT checkpoint. \rev{Panel (e)} jointly trains the two predictions
through their combined guided field. Training uses 512 rollouts per update,
eight rollout steps, learning rate 0.002, and ten outer updates; evaluation
uses 32 steps.
Hit denotes $d_{\mathrm{target}}\leq0.04$, and on-tree denotes
$d_{\mathrm{tree}}\leq0.02$. Reward is
$\exp[-(d_{\mathrm{target}}/0.55)^2]$, so target proximity and geometric
fidelity are separate measurements. Metrics use all retained endpoints
without clipping to the display window.

\input{tables/toy_nft_four_way_metrics}

\subsection{\texorpdfstring{\rev{Isolated Effect of Center Error}}{Isolated Effect of Center Error}}
\label{app:toy-center-bias}
{\revon
Equation~\ref{eq:center-error} predicts that a center error
$e=C-\bar v_{q_{\rm CFG}}$ adds the center-induced bias
$[(1-2\alpha)/\beta]e$ to the reward-driven displacement. We isolate the two
terms, with $\beta=1$, in the CFG sampler of the tree model above
(class-A width-64 conditional model, class-AB width-32 guide, $w=4$) by
applying each displacement directly, without training a network.
A single rollout of 65{,}536 endpoints is sampled from the initial field
with 32 steps. The re-noised endpoint population is treated as a Gaussian
mixture, from which $\bar v_{q_{\rm CFG}}$, $\alpha$, and the
reward-driven displacement are evaluated exactly on a $448\times448$ grid
at every sampling noise level $\sigma\ge0.02$. This rollout is reused in
each of ten rounds, in which the selected displacement is added pointwise
to the sampler field, so the settings in
Figure~\ref{fig:toy-guidance-nft-center} differ only in the regression
center. We compare the matched center $C=\bar v_{q_{\rm CFG}}$, for which
the update reduces to the reward-driven displacement, with an unmatched
center, the frozen conditional prediction $C=v_c$. For the unmatched center
we apply either the center-induced bias alone or the full displacement
$d^*$ of Equation~\ref{eq:center-error}. Hit is defined as above; cells,
the effective number of occupied cells, is the exponential of the entropy
of the endpoint histogram with cell width 0.02.\par}

\input{experiment_appendix}

%% file: tables/toy_nft_four_way_metrics.tex
\begin{table}[!ht]
\centering
\small
\setlength{\tabcolsep}{9pt}
\caption{\textbf{Endpoint metrics for the toy experiment in Appendix~\ref{app:toy-guidance-nft}.} Figure~\ref{fig:toy-guidance-nft} displays the same five configurations.}
\label{tab:toy-guidance-nft}
\begin{tabular}{@{}lrrr@{}}
\toprule
\textbf{Method} & \textbf{Reward} $\uparrow$ & \textbf{Hit} $\uparrow$ & \textbf{On-tree} $\uparrow$ \\
\midrule
Base & 0.755 & 0.254 & 0.848 \\
CFG & 0.721 & 0.347 & \textbf{0.946} \\
NFT & 0.956 & 0.630 & 0.805 \\
NFT + CFG & 0.985 & 0.851 & 0.715 \\
Guided-field NFT & \textbf{0.996} & \textbf{0.971} & 0.907 \\
\bottomrule
\end{tabular}
\end{table}

%% file: experiment_appendix.tex
\section{Training and Evaluation Protocols}
\label{app:experimental-protocol}

We distinguish the rollout distribution, the center in the NFT loss, and
the prediction that receives its gradient. Write
$g_{\theta,w}=w v_{\theta,c}+(1-w)v_{\theta,\varnothing}$.
``Conditional update'' identifies the prediction in the loss; shared
parameters can also change the empty prediction. Only an explicitly frozen
pretrained branch remains fixed at inference.

\begin{table}[ht]
\centering\small
\setlength{\tabcolsep}{3pt}
\renewcommand{\arraystretch}{1.12}
\begin{tabular}{@{}p{0.29\linewidth}p{0.13\linewidth}p{0.23\linewidth}p{0.27\linewidth}@{}}
\toprule
Method & Rollout law & Loss center & Loss prediction \\
\midrule
\multicolumn{4}{@{}l}{\textit{SD3.5 and SiT: explicit CFG}} \\
NFT & $q_1$ & Conditional output & Conditional output \\
NFT + pretrained prior & $q_1$ & Conditional output & Conditional output \\
NFT (old center) & $q_w$ & Conditional output & Conditional output \\
NFT (analytic center) & $q_w$ & Analytic CFG field & Conditional output \\
GuidanceNFT & $q_w$ & Fitted center & Conditional output \\
\midrule
\multicolumn{4}{@{}l}{\textit{Model Guidance: intrinsic guidance}} \\
NFT & $q_w$ & Native guided field & Native guided field \\
GuidanceNFT & $q_w$ & Fitted center & Native guided field \\
\midrule
\multicolumn{4}{@{}l}{\textit{ELF-B reasoning: native SCCFG 2}} \\
NFT & $q_w$ & Native guided field & Native guided field \\
GuidanceNFT & $q_w$ & Fitted center & Native guided field \\
CEDR & $q_w$ & Reconstructed base prediction & \shortstack[l]{Native guided field with\\detached correction} \\
\bottomrule
\end{tabular}
\caption{Rollout law, loss center, and loss prediction.
$q_1$ denotes unguided rollouts and $q_w$ guided rollouts, including
intrinsic guidance with external CFG 1. SD3.5 uses $w=4$, SiT uses
$w=1.5$, and ELF-B reasoning uses native SCCFG 2.
Conditional output denotes $v_{\theta,c}$; the native guided field is
the model's built-in guided prediction. Loss centers use detached old outputs or a
fitted center. For CEDR, the reconstructed center and loss prediction
apply its detached SCCFG correction. NFT + pretrained prior~\citep{phunyaphibarn2026unconditional} changes only the
empty branch at inference.}
\label{tab:experiment-contract}
\end{table}

The inference-only
pretrained-empty control uses
$w v_{\theta,c}+(1-w)v_{{\rm pre},\varnothing}$ on the NFT checkpoint;
it does not require an additional policy-training run.
For Qwen, the local NFT and GuidanceNFT runs disable velocity reconstruction.
Appendix~\ref{app:lfrm-nft-implementation} separately specifies
CEDR's detached SCCFG correction.

\paragraph{SD3.5 settings.}
The center comparison uses five rollout steps, $512\times512$ resolution,
48 prompt groups with 24 images each, one policy update per iteration,
$\beta=1$, and reference coefficient $10^{-4}$.
All three runs update the conditional prediction. Their corrections are
$\mathcal C=0$, $\operatorname{sg}(v_4^{\rm old}-v_c^{\rm old})$,
and $\operatorname{sg}(v_\phi-v_c^{\rm old})$, respectively.
The learned center is persistent, with one fitting epoch per refresh,
learning rate $3\times10^{-4}$, gradient clipping at 1, and EMA decay 0.9.
The loss uses clean-sample prediction space and detached branch-MAE
normalization. These runs share the stated rollout protocol.

The update-200 GenEval, PickScore, and Learned3 runs use ten-step
training rollouts. Learned3 update 100 is a separate training run.
NFT and GuidanceNFT omit conditioning dropout. Learned3 uses $\beta=0.1$
and GenEval $\beta=1$; both use reference coefficient $10^{-4}$.
Learned3 optimizes PickScore/26 + CLIPScore + HPSv2.1;
reported PickScore is unscaled.

Scheduled PickScore assessments use 2,048 Pick-a-Pic test prompts,
40 steps, and a fixed evaluation seed. Logged scores are multiplied by 26 to recover
raw PickScore. These assessments use held-out prompts, separate from
rollout rewards and DrawBench. The training splits contain 50,000
GenEval metadata rows and 25,432 Learned3 prompts, separate from
DrawBench. OCR uses 1,018
evaluation examples per seed and is absent from these training objectives.

\paragraph{Reward and loss conventions.}
PickScore/Pick-a-Pic~\citep{kirstain2023pickapic}, HPSv2~\citep{wu2023hpsv2},
and ImageReward~\citep{xu2023imagereward} provide learned preference scores.
HPS uses the v2.1 checkpoint. The CLIPScore column follows NFT's scorer:
the paired image--text logit divided by 100, with CLIP ViT-L/14~\citep{radford2021clip} as its
source-code default. It is not the original rescaled CLIPScore definition
of \citet{hessel2021clipscore}.
SD3.5 uses AdamW with learning rate $3\times10^{-4}$, coefficients
$(0.9,0.999)$, weight decay $10^{-4}$, $\epsilon=10^{-8}$, and gradient
clipping at 1. Training uses FP16, two H200 GPUs, and LoRA~\citep{hu2022lora}; the retained
Learned3 checkpoint specifies rank 32, scaling parameter 64, and dropout 0.
The conditional microbatch is 24 per GPU with accumulation 24.
The sampler is deterministic DPM2. The source selects a fraction 0.99 of
the rollout time grid for NFT training.

Let $s_i$ be the summed raw task score, $\bar s_{c_i}$ its prompt-group
mean in the current collected batch, and $\sigma_B$ the standard deviation
across that batch. The archived SD3.5 mapping is
\begin{equation}
A_i=\frac{s_i-\bar s_{c_i}}{\sigma_B+10^{-4}},\qquad
R_i=\frac12+\frac{\operatorname{clip}(A_i,-5,5)}{10}.
\end{equation}
Prompt statistics are cleared after each outer update. The NFT branch losses
receive weights $R_i$ and $1-R_i$, with an overall multiplier $5/\beta$.
In clean-sample prediction space, a velocity residual is multiplied by the noise
time. Its squared residual is divided by its detached per-example mean
absolute residual, lower-bounded by $10^{-5}$. The reference term penalizes
a prediction difference.

Held-out GenEval aggregates retain valid negative rewards, including
$-10$. Historical online reward displays exclude $-10$, although the
advantage calculation uses the full reward array.
Figure~\ref{fig:explicit-guidance-diagnosis}(b) retains this recorded
online metric; held-out evaluation uses the full aggregation.

\paragraph{ELF-B settings.}
OpenWebText training uses 384 endpoints per outer update, eight time draws
per endpoint, 100 updates, learning rate $3\times10^{-5}$, $\beta=1$,
reference coefficient 0.1, decoder cross-entropy coefficient 0.25, and an
adaptive entropy-floor term targeting 5.1 nats. The quality evaluator is
GPT-2 large. Native SCCFG changes the model's self-conditioning field;
external CFG is 1 throughout the reported sweep. The comparison matches
policy-update counts, and learned-center training adds estimator updates.
Every sweep point uses 1,024 samples, a shared evaluation seed, 32 SDE steps, and
$\gamma=1.5$. The separate repetition diagnostic uses 1,000 samples, so
its PPL and entropy numbers are not interchangeable with the sweep values.

Figure~\ref{fig:sccfg-main-sweep} evaluates the pretrained model,
CEDR-style reconstruction trained with SCCFG-1 rollouts, and GuidanceNFT
trained with SCCFG-3 rollouts. Reconstruction samples guidance inputs
for its regression loss independently of rollout guidance. Each
fine-tuned checkpoint is at update 100, and each curve varies evaluation
SCCFG. The repetition diagnostic separately compares native-field NFT
and GuidanceNFT trained at SCCFG 3. These are comparisons of the stated
training configurations, including their rollout guidance.

\paragraph{SiT and Model Guidance settings.}
Both use class-name PickScore, 1,152 endpoints per update, nine time draws
per endpoint, ten-step SDE rollouts, 400 policy updates, learning rate
$3\times10^{-6}$, $\beta=1$, and reference coefficient $10^{-4}$. SiT
NFT trains at CFG 1; GuidanceNFT trains at CFG 1.5. Model Guidance uses intrinsic guidance with external
CFG 1 for both methods. The fitted-center EMA decay is 0.9.
Evaluation uses three seeds, each with two samples for every ImageNet
class. Center training adds computation but no estimator evaluation at
inference. Missing MG log entries remain gaps; late averages use the
available final updates.

\paragraph{Display conventions for training curves.}
In Figure~\ref{fig:explicit-guidance-diagnosis}(a), ``NFT + base'' reuses
the frozen pretrained unconditional prediction following
\citet{phunyaphibarn2026unconditional}; bars show sample std over three
evaluation seeds. Panel (c) compares each training center with an FM100
reference fitted to that run's own rollout endpoints; shading spans the
mean to one prompt std above it at the re-noising level $\sigma=0.5$.

Figure~\ref{fig:intrinsic-guidance-dynamics}(a) uses a 30-update trailing
mean for both methods, with missing updates kept as gaps. Panel (b) uses
15-update centered means over 127 shared updates; both panels retain faint
raw traces. Panel (c) connects measured checkpoint means with
straight line segments and marks each measured checkpoint. The same solid
NFT and dashed GuidanceNFT styles apply throughout.
The 32-step synchronous evaluation in this panel differs from the 64-step
asynchronous protocol in Table~\ref{tab:qwen-full1319-nft-comparison}.
Figure~\ref{fig:explicit-guidance-diagnosis}(b) uses a Gaussian display
filter with a five-update standard deviation and reflected boundaries.

\paragraph{Selection and uncertainty.}
Table~\ref{tab:checkpoint-selection} records checkpoint selection.
SD3.5 uses fixed checkpoints. SiT NFT selects the best PickScore among
updates 100--400 at intervals of 50, choosing update 200 at both CFGs;
MG NFT selects update 200 over 400 from the earlier fixed-seed assessment.
Both GuidanceNFT image comparisons use update 400. Qwen NFT selects the best
Pass@1 over updates 50--400 at intervals of 50, while GuidanceNFT uses update
500.

\input{tables/problem2_checkpoint_selection}

Standard deviations use the sample denominator $n-1$. Evaluation-seed
std uses two seeds for fine-tuned SD3.5 rows and three for SiT/MG;
Learned3 sums reward components within each seed before computing std.
Paired differences average all images or answers within each prompt or
question, then report std across these units. GenEval task-macro
comparisons instead use the six task means. Qwen's main table reports std across
question-bootstrap means. Fixed-checkpoint paired comparisons appear in
Table~\ref{tab:revision-fixed-ci}.

\input{tables/revision_fixed_checkpoint_ci}

\section{Additional SD3.5 Results}
\label{app:sd35-tables}

\paragraph{Metrics and uncertainty.}
Table~\ref{tab:sd35-cfg4-update200} evaluates GenEval-, OCR-, and
Learned3-trained checkpoints at update 200 using ten-step training
rollouts and 40-step inference. NFT trains at CFG 1 and is evaluated at
CFG 1/4; GuidanceNFT trains and is evaluated at CFG 4. Train reward is the
optimized objective on its on-domain test set. The seven evaluator
columns use separately generated samples; gray cells mark evaluators
included in training. All scores are higher-is-better, and bold includes
ties at the displayed precision.

Every fine-tuned row averages two evaluation seeds equally; pretrained rows
use one seed. Table~\ref{tab:sd35-table1-seed-sd} reports the corresponding
means and sample stds. Each seed uses 2,212 GenEval
images, 1,018 OCR images, 2,048 Pick-a-Pic images for Learned3 Train
reward, and 1,000 DrawBench images for the five general-quality metrics.
Learned3 Train reward sums raw PickScore/26, HPSv2.1, and CLIPScore
within each seed.

\input{tables/sd35_table1_seed_sd}

GenEval Train reward uses the native dense reward, which grants partial
credit and retains valid negative values. The GenEval evaluator column
uses image-micro averages of saved success flags. The benchmark's
alternative task-macro aggregation weights its six tasks equally:
on one evaluation seed it gives 0.8547/0.5762 for NFT at CFG 1/4 and 0.8815 for
GuidanceNFT at CFG 4. We use saved \texttt{accuracy} flags from the official
all-condition success rule~\citep{ghosh2023geneval}; the local
\texttt{strict\_accuracy} field instead uses an exact-count rule.
The detector and color-model inference are those of the saved evaluation.
An evaluator excluded from training tests reward transfer; prompt
holdout is specified separately by each evaluation suite.

\paragraph{Where the GenEval gains arise.}
The task breakdown in Table~\ref{tab:problem2-geneval-tasks} shows broad
improvement at CFG 4. Among 553 paired prompts, GuidanceNFT increases the
number of successful images for 285, decreases it for 25, and leaves
243 unchanged. Across image pairs, 754 failures become successes and
64 successes become failures. Each prompt contributes its four-image
mean to the within-task std of paired differences.

\input{tables/problem2_geneval_tasks}

\subsection{Qualitative comparisons}
\label{app:sd35-qualitative}

Figure~\ref{fig:sd35-learned3-ocr-qualitative} combines \rev{four} Learned3
DrawBench samples at update 300 with \rev{four} OCR
online-evaluation previews at update 200. Learned3
candidates were ranked by GuidanceNFT's combined reward,
$\mathrm{PickScore}/26+\mathrm{HPSv2.1}+\mathrm{CLIPScore}$, among paired
samples for which both NFT settings scored below their respective medians
and GuidanceNFT exceeded both. Repeated prompts were removed from the displayed
shortlist; the first three distinct prompts \rev{and one further shortlisted
prompt} were selected for the figure.
The OCR cases were selected from the saved preview grid. Both groups use
40 inference steps, with noise paired across methods within each column.
These are selected qualitative comparisons; aggregate performance is
reported in Table~\ref{tab:sd35-cfg4-update200}.

\begin{figure}[t]
\centering
\includegraphics[width=\linewidth]{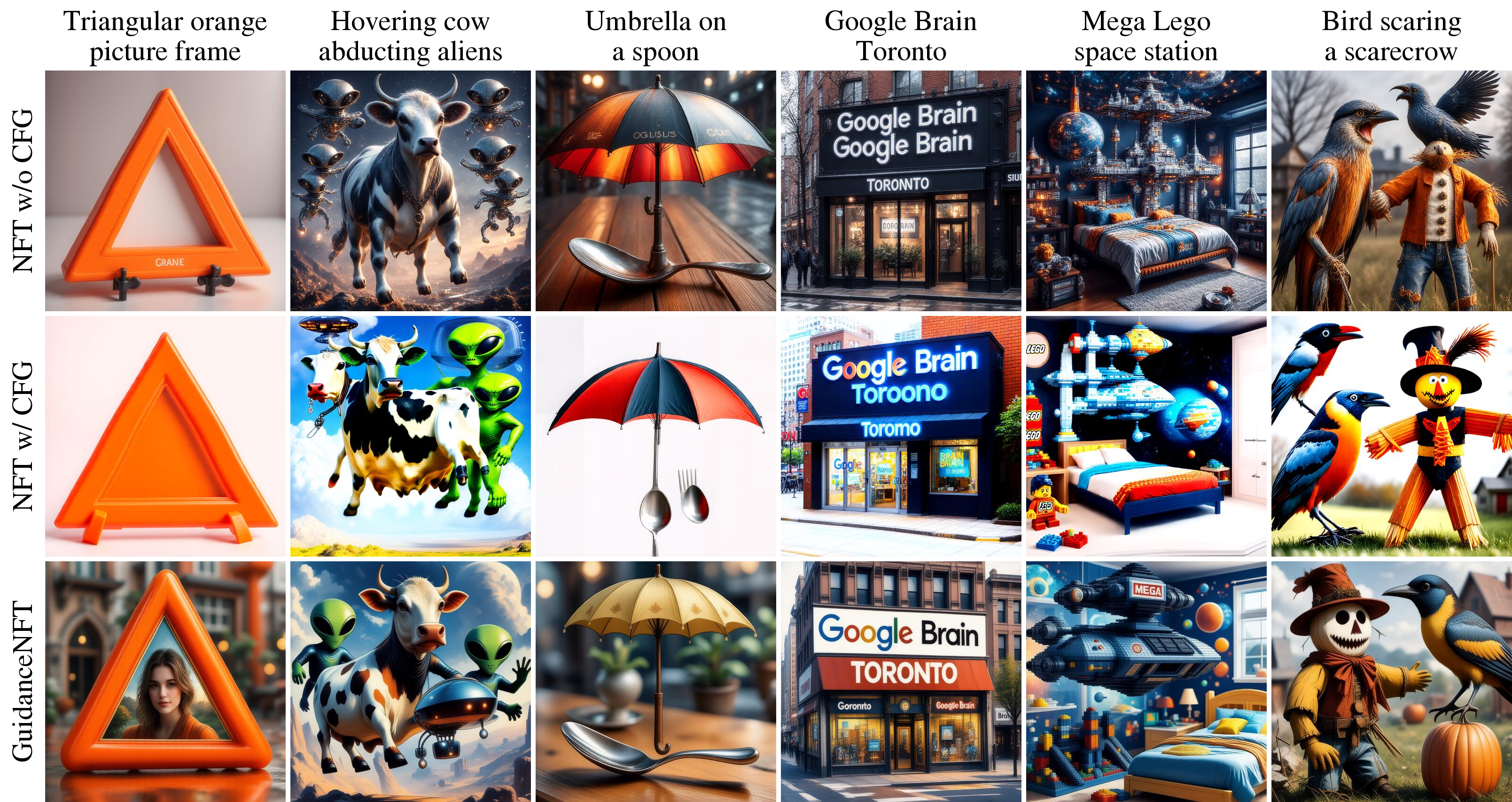}
\caption{\textbf{Six-prompt Learned3 comparison.}
Selected SD3.5 DrawBench samples at update 300, using 40 inference steps
at $512\times512$ resolution. Rows show NFT at CFG 1, NFT at CFG 4,
and GuidanceNFT at CFG 4.
Headings summarize each prompt; noise is paired across methods.}
\label{fig:sd35-learned3-six-column}
\end{figure}

\begin{figure}[t]
\centering
\includegraphics[width=\linewidth]{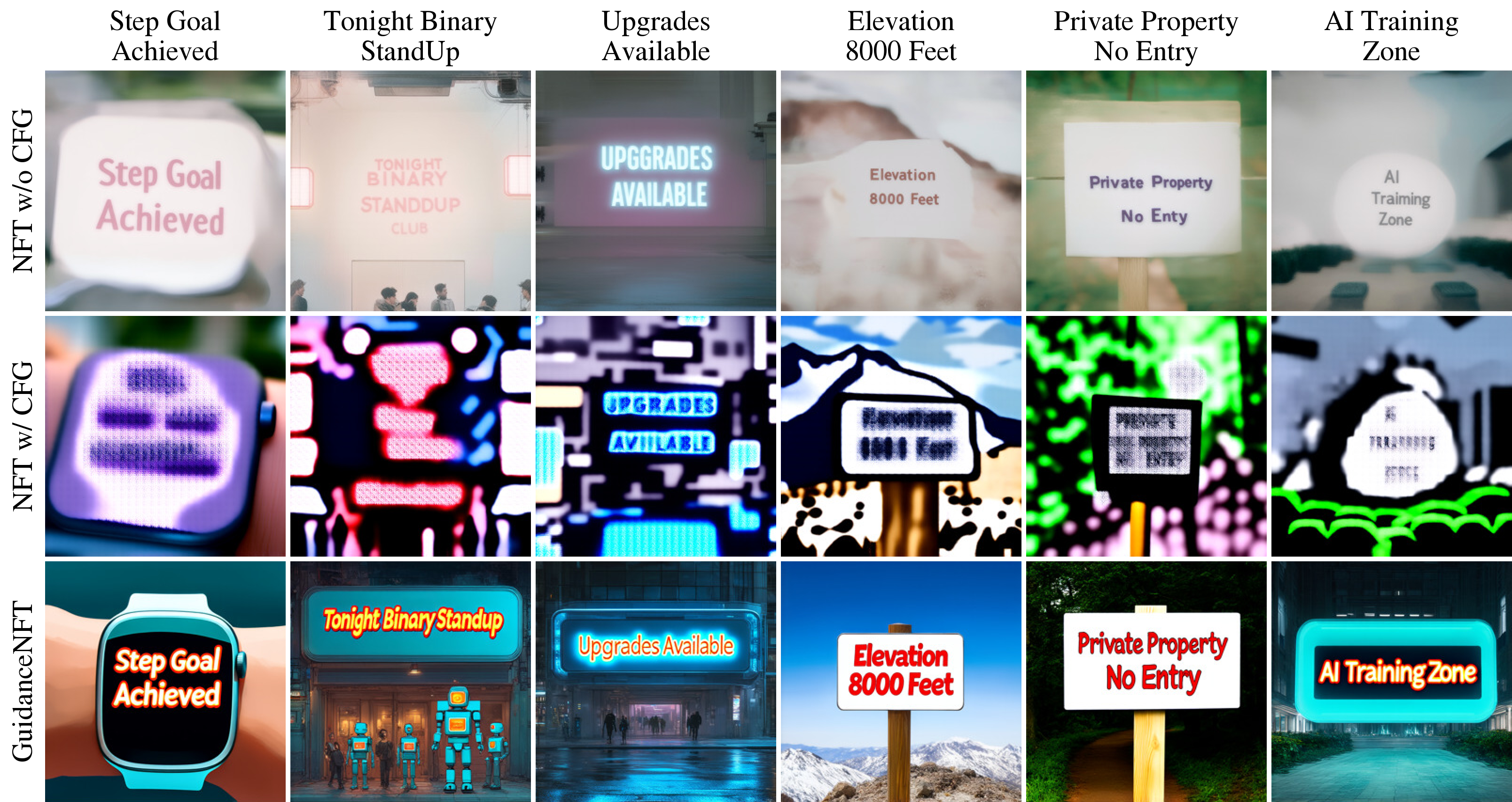}
\caption{\textbf{Six-prompt OCR comparison.}
Selected SD3.5 online-evaluation previews at update 200, using 40 inference
steps. Rows show NFT at CFG 1, NFT at CFG 4, and GuidanceNFT at CFG 4.
Headings give the target text.
The saved previews have native resolution $256\times256$;
noise is paired across methods within each column.}
\label{fig:sd35-ocr-six-column}
\end{figure}

Figures~\ref{fig:sd35-learned3-six-column} and
\ref{fig:sd35-ocr-six-column} extend these comparisons to six prompts
per objective. The Learned3 examples compare object, spatial, and text
alignment; the OCR examples compare target-text rendering and scene content.
These selected examples use paired noise within each column.

The full prompts for Figure~\ref{fig:sd35-learned3-ocr-qualitative},
from left to right, are reproduced verbatim below. Figure headings are
short summaries rather than the full generation instructions.
Learned3 images are $512\times512$ outputs; OCR images are saved
$256\times256$ previews, enlarged by nearest-neighbor pixel replication.
\begin{enumerate}
\item A triangular orange picture frame. An orange picture frame in the shape of a triangle.
\item Hovering cow abducting aliens.
\item An umbrella on top of a spoon.
\item \rev{A bird scaring a scarecrow.}
\item A close-up of a sleek smartwatch on a wrist, the screen displaying "Step Goal Achieved" with a celebratory animation, set against a blurred cityscape at dusk, capturing the moment of accomplishment.
\item A futuristic urban street at night, a cyborg shop sign glowing "Upgrades Available" in neon blue, reflecting off wet pavements, with cybernetic pedestrians and sleek, futuristic buildings in the background.
\item A high-altitude mountain summit with a wooden signpost clearly marked "Elevation 8000 Feet", surrounded by rocky terrain and a backdrop of distant, snow-capped peaks under a clear blue sky.
\item \rev{A realistic photo of a tech campus courtyard at night, featuring a glowing "AI Training Zone" hologram floating in the center, surrounded by futuristic buildings and greenery, with soft ambient lighting enhancing the futuristic atmosphere.}
\end{enumerate}

\begin{figure}[t]
\centering
\includegraphics[width=\linewidth]{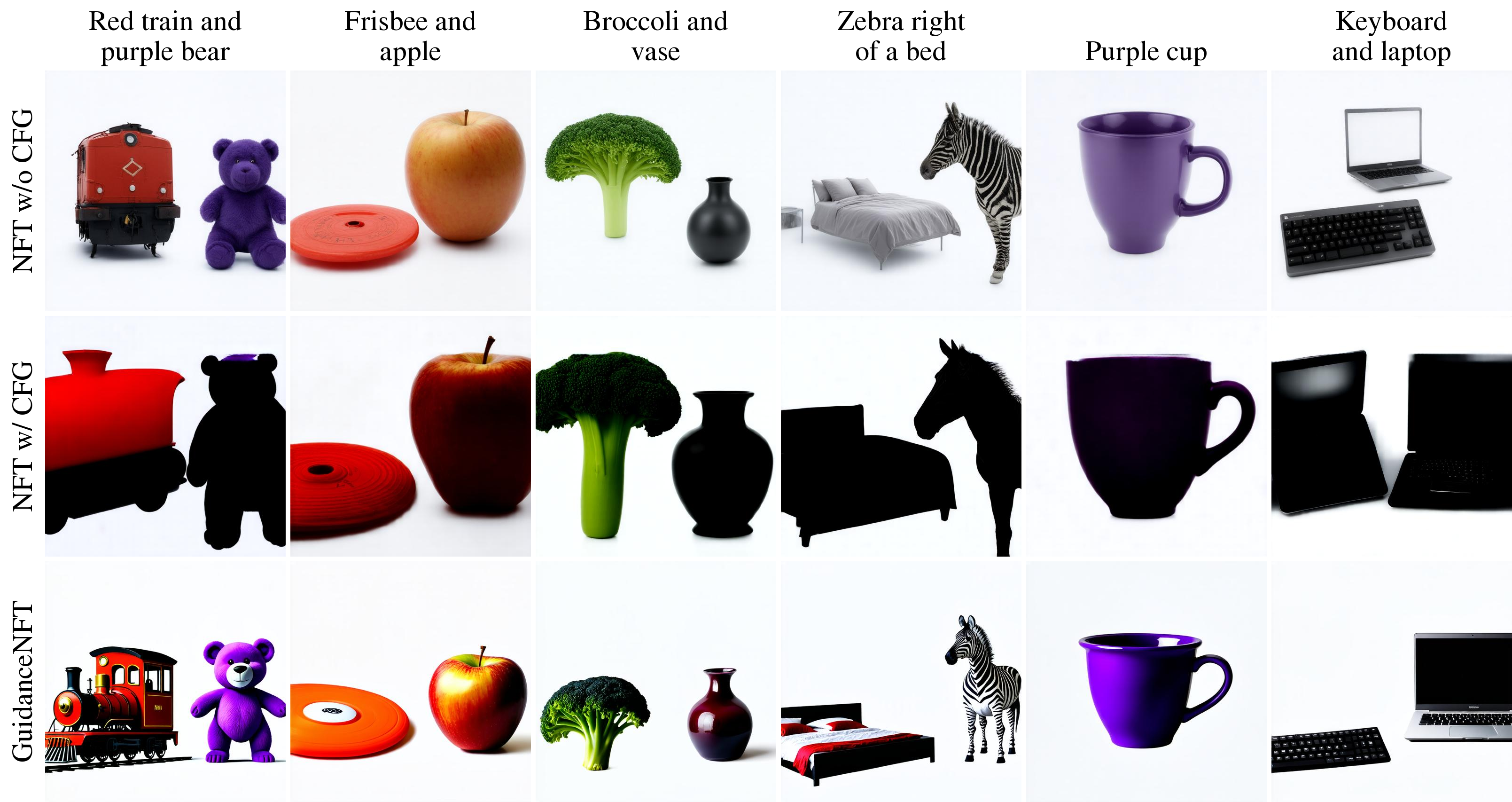}
\caption{\textbf{Center correction preserves color and shading under CFG.}
Selected GenEval-trained SD3.5 samples at update 200, with NFT at CFG 1
(top), NFT at CFG 4 (middle), and GuidanceNFT at CFG 4 (bottom).
Columns pair latent noise across methods, using 40 steps and
$512\times512$ resolution. Headings abbreviate the prompts.}
\label{fig:sd35-update200-diagnosis}
\end{figure}

Figure~\ref{fig:sd35-update200-diagnosis} displays selected GenEval-specialist
examples at update 200. Matching the seed controls the noise draw within a
prompt. NFT CFG-4 samples show less texture and shading in these examples;
the GuidanceNFT samples remain stylized. The six paired cases complement the aggregate GenEval comparison.

\FloatBarrier
\section{Empirical Center Diagnostics and Training Costs}
\label{app:center-diagnostics}

\paragraph{Diagnostic design.}
We test whether a separate predictor better estimates the mean velocity
of guided endpoints. For each frozen checkpoint, an auxiliary FM model
is fitted for exactly 100 AdamW updates on 384 endpoints, with 96 held
out: eight fitting and two held-out endpoints for each of 48 classes
(MG) or category-balanced prompts (SD3.5). Classes and prompts are shared
between fitting and evaluation. The auxiliary EMA, denoted
$C_{\rm FM100}$, is a finite-fit reference. Each comparison uses identical
held-out endpoints, noise levels, and Gaussian draws for both fields.
We report two complementary statistics:
\begin{equation}
D_q(F)=L_q(F)-L_q(C_{\rm FM100}),\qquad
G_q(F)=\mathbb E_q\|F-C_{\rm FM100}\|^2/d,
\label{eq:empirical-anchor-risk-gap}
\end{equation}
where $d$ is the field dimension. The paired risk difference $D_q$
cancels irreducible conditional variance and compares center MSEs
(Equation~\ref{eq:paired-fm-risk}); $G_q$ measures field disagreement.
Standard deviations use the 48
class or prompt means, averaging paired endpoints within each unit.

\paragraph{Model Guidance.}
NFT at update 200 and GuidanceNFT at update 400 each generate their own
population with ten-step Euler--Maruyama rollouts and external CFG 1.
Each auxiliary predictor starts from the corresponding native policy,
with learning rate $3\times10^{-6}$, effective batch 96, and EMA decay
0.9. Evaluation averages 11 fixed noise levels from 0.1 to 0.9 equally.

The native-minus-FM100 risk is $0.6969\pm0.7250$ on NFT endpoints and
$0.0712\pm0.0201$ on GuidanceNFT endpoints (class std). On the latter
population, the saved GuidanceNFT center has risk difference
$-0.00715\pm0.00396$, outperforming FM100. Directly comparing the native
field with the saved center gives $0.07835\pm0.02287$, supporting
separate center estimation on the same rollout population.

\begin{figure}[tbp]
\centering
\includegraphics[width=\linewidth]{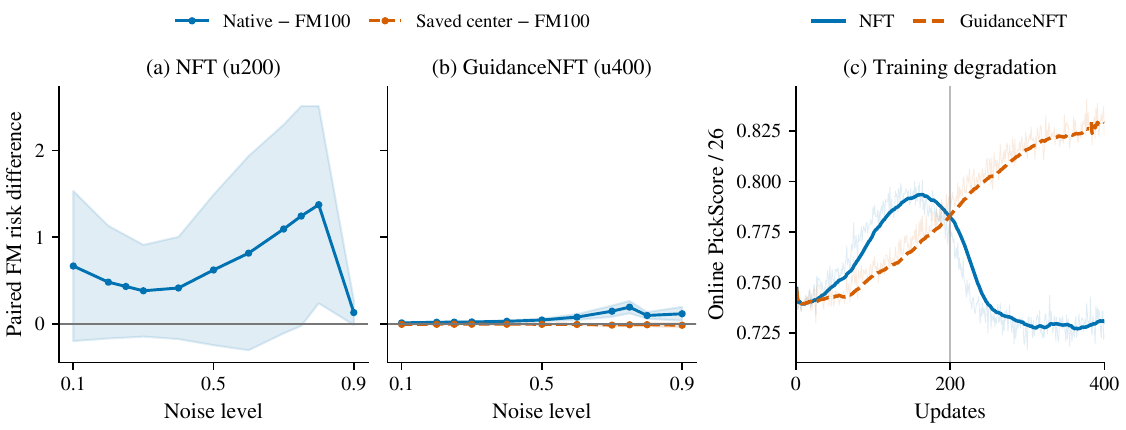}
\caption{\textbf{A fitted center lowers held-out flow-matching risk on Model Guidance.}
(a--b) Paired risk differences relative to each checkpoint's FM100 EMA:
native field (blue) and saved GuidanceNFT center (orange).
Bands show mean $\pm$ sample std over 48 class means.
(c) Online rewards, with 30-update trailing means and faint raw traces;
the vertical line marks NFT update 200. The two checkpoint populations
are evaluated separately.}
\label{fig:mg-anchor-risk}
\end{figure}

\paragraph{SD3.5.}
We evaluate the three five-step CFG-4 runs in
Figure~\ref{fig:explicit-guidance-diagnosis}(b) at updates 0, 50, and 100.
Each saved old-policy adapter generates its own deterministic DPM2
endpoints. The tested centers are the old conditional field, the
analytic CFG-4 field, and the saved center EMA, respectively.
Prompts and initial-noise seeds are paired across runs; each run retains
its own endpoints and independently fitted FM100 reference.
The auxiliary models start from their respective old conditional fields
and use learning rate $3\times10^{-4}$, effective batch 8, zero weight
decay, gradient clipping at 1, and dynamic EMA capped at 0.9.
Fitting draws fresh Gaussian noise and uniform continuous times while
the policy and saved training center remain frozen.

Evaluation uses $\sigma\in\{0.1,0.2,\ldots,0.9\}$ with equal weights.
At initialization the learned center equals the uncorrected conditional
field; differences between their initial diagnostic gaps arise from
separate auxiliary fits. At $\sigma=0.5$, between updates 50 and 100,
field disagreement grows from 0.156 to 0.219 without correction and
from 5.43 to 148.06 with analytic correction, while GuidanceNFT remains near
0.008 (Figure~\ref{fig:explicit-guidance-diagnosis}(c)). The corresponding
noise-averaged gaps change from 0.1999 to 0.2922, from 5.6244 to 127.3269,
and from 0.01135 to 0.01183. Learned correction thus maintains agreement
with its rollout reference as the two alternatives diverge.

\begin{figure}[tbp]
\centering
\includegraphics[width=\linewidth]{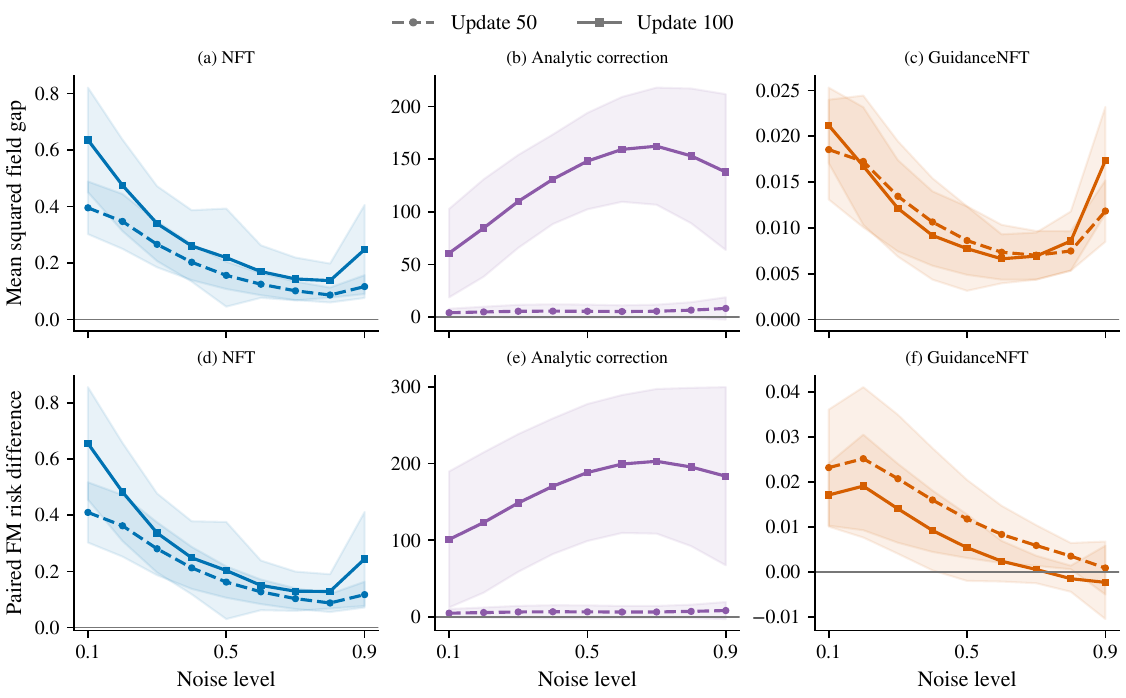}
\caption{\textbf{Center correction controls the growing mismatch in SD3.5.}
Columns compare no correction, analytic correction, and learned correction.
Each run uses its own endpoints and FM100 EMA at update 50 (dashed)
or 100 (solid). Top: squared field disagreement $G_q$.
Bottom: paired FM-risk difference $D_q$, which can be negative.
Bands show mean $\pm$ sample std across 48 prompt means.
All policies and saved training centers are frozen during diagnosis.}
\label{fig:sd35-anchor-risk}
\end{figure}

The same runs show a corresponding separation in reward trajectories.
First-ten to last-ten mean rewards over updates 0--99 fall from 0.3085
to 0.0186 with the analytic center and from 0.3251 to 0.0953 with the
old conditional center, but rise from 0.3198 to 0.4977 with learned
correction. A ten-update trailing mean falls below 0.1 at updates 41
and 97 for the two failing runs, respectively. These trace summaries
support the association between center mismatch and degradation.
The exact displacement error also weights center error by
$(1-2\alpha)/\beta$ (Equation~\ref{eq:center-error}).

\subsection{\texorpdfstring{\rev{Perturbing the Learned Center}}{Perturbing the Learned Center}}
\label{app:center-perturbation}

\begin{figure}[!ht]
\centering
\includegraphics[width=0.8\linewidth]{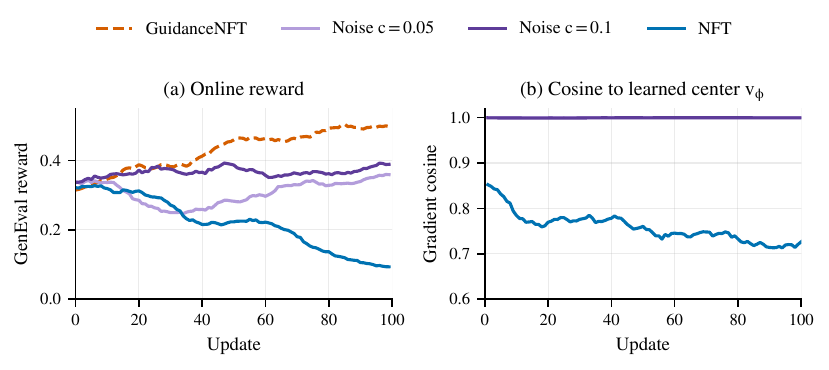}
\caption{{\revon\textbf{Training with a noisy center on SD3.5-M.}
Left: online GenEval reward. Right: cosine between the policy-loss gradient
under the training center and under the learned center $v_\phi$. Curves
are smoothed with a centered moving average.}}
\label{fig:center-perturb-noise}
\end{figure}

\begin{figure}[!ht]
\centering
\includegraphics[width=\linewidth]{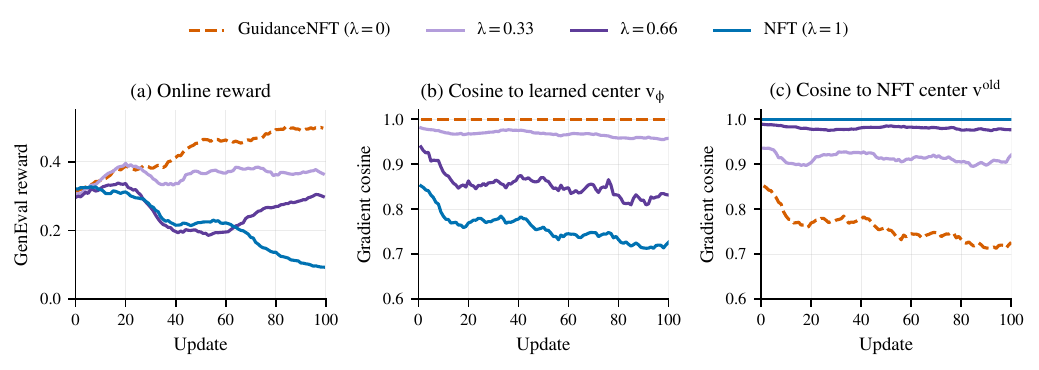}
\caption{{\revon\textbf{Training with an interpolated center on SD3.5-M.}
Left: online GenEval reward. Middle and right: cosine between the
policy-loss gradient under the training center and under the learned
center $v_\phi$ or NFT's center $v^{\rm old}$. Curves are smoothed with a
centered moving average.}}
\label{fig:center-perturb-lambda}
\end{figure}

{\revon We test two ways to perturb the learned center. Both use the five-step
CFG-4 GenEval setting of Figure~\ref{fig:explicit-guidance-diagnosis}(b)
and train for 100 updates; GenEval is evaluated with seed 42 and 40
sampling steps. The gradient cosine compares, on the same batch, the
policy-loss gradient under the training center with that under the
learned center $v_\phi$, averaged over updates.

\paragraph{Noise.}
We add Gaussian noise to the center, with standard deviation $c$ times its
per-example root-mean-square. The perturbed gradients remain closely
aligned with the gradient under $v_\phi$, but both noise levels reduce
the final reward (Table~\ref{tab:center-perturb-noise} and
Figure~\ref{fig:center-perturb-noise}).

\paragraph{Interpolated center.}
We also control the scale of the center mismatch error by using
$\lambda v^{\rm old}+(1-\lambda)v_\phi$ as the center, where $\lambda=0$
is GuidanceNFT and $\lambda=1$ is NFT. As $\lambda$ decreases, the
gradient aligns more closely with the gradient under $v_\phi$, and the
reward improves (Table~\ref{tab:center-perturb-lambda}
and Figure~\ref{fig:center-perturb-lambda}).\par}

\begin{table}[!ht]
\centering
\small
\setlength{\tabcolsep}{6pt}
\caption{{\revon\textbf{Noise added to the learned center on SD3.5-M.}
Five-step CFG-4 rollouts, GenEval objective. The gradient cosine compares
the policy-loss gradient under the training center with that under
$v_\phi$ on the same batch, averaged over 100 updates. GenEval is
evaluated at update 100 with seed 42 and 40 sampling steps.}}
\label{tab:center-perturb-noise}
\begin{tabular}{@{}lcccc@{}}
\toprule
& GuidanceNFT & noise $c=0.05$ & noise $c=0.1$ & NFT \\
\midrule
Gradient cosine & 1 & 1.000 & 0.999 & 0.759 \\
GenEval@100, CFG 1 & 0.7534 & 0.6527 & 0.7113 & 0.5616 \\
GenEval@100, CFG 4 & 0.8173 & 0.7141 & 0.7691 & 0.6039 \\
\bottomrule
\end{tabular}
\end{table}

\begin{table}[!ht]
\centering
\small
\setlength{\tabcolsep}{6pt}
\caption{{\revon\textbf{Interpolating between the NFT and learned centers
on SD3.5-M.} The center is $\lambda v^{\rm old}+(1-\lambda)v_\phi$, where
$\lambda=0$ is GuidanceNFT and $\lambda=1$ is NFT. Settings and metrics
follow Table~\ref{tab:center-perturb-noise}.}}
\label{tab:center-perturb-lambda}
\begin{tabular}{@{}lcccc@{}}
\toprule
& $\lambda=0$ & $\lambda=0.33$ & $\lambda=0.66$ & $\lambda=1$ \\
\midrule
Gradient cosine & 1 & 0.967 & 0.854 & 0.759 \\
GenEval@100, CFG 1 & 0.7534 & 0.7104 & 0.6457 & 0.5616 \\
GenEval@100, CFG 4 & 0.8173 & 0.7567 & 0.7261 & 0.6039 \\
\bottomrule
\end{tabular}
\end{table}

\subsection{\texorpdfstring{\rev{Fixed Rollout Population}}{Fixed Rollout Population}}
\label{app:fixed-rollout}

{\revon In the diagnostics above, the rollout populations of the compared
runs evolve separately, so a shift in the rollouts could cause the center
mismatch rather than result from it. We therefore fix the rollout
population. On SiT-XL/2 with PickScore, every update draws its CFG-1.5
rollouts from the frozen pretrained model rather than from the current
policy, so both methods train on the same, unchanging distribution.
Standard NFT uses the old conditional prediction as its center; GuidanceNFT
fits its center to these rollouts.
Other settings follow the SiT runs, and evaluation uses the ten-step
benchmark of Table~\ref{tab:sit-imagenet-checkpoints}.

Standard NFT improves until update 100--150 and then declines below the
pretrained model (PickScore/26 of 0.7250 with CFG 1 and 0.7418 with
CFG 1.5), whereas GuidanceNFT keeps improving
(Table~\ref{tab:fixed-rollout}). Since the rollout distribution is
identical for both methods and does not change during training, the
decline of NFT is not caused by a shift in its own rollouts; it appears
only with the mismatched center.\par}

\begin{table}[!ht]
\centering
\small
\setlength{\tabcolsep}{6pt}
\caption{{\revon\textbf{Training on a fixed rollout population (SiT-XL/2).}
Rollouts are drawn with CFG 1.5 from the frozen pretrained model at every
update. PickScore/26 under the ten-step benchmark (6,000 images over three
evaluation seeds), shown as the mean with $\pm$ sample std across the three
evaluation-seed means beneath it; the pretrained model scores 0.7250 (CFG 1) and 0.7418
(CFG 1.5).}}
\label{tab:fixed-rollout}
\begin{tabular}{@{}lcccccc@{}}
\toprule
& \multicolumn{3}{c}{Eval CFG 1} & \multicolumn{3}{c}{Eval CFG 1.5} \\
\cmidrule(lr){2-4}\cmidrule(l){5-7}
Method & Update 100 & Update 150 & Update 200 & Update 100 & Update 150 & Update 200 \\
\midrule
NFT & \shortstack{$0.7466$\\$\pm 0.0006$} & \shortstack{$0.7428$\\$\pm 0.0006$} & \shortstack{$0.7139$\\$\pm 0.0009$} & \shortstack{$0.7631$\\$\pm 0.0005$} & \shortstack{$0.7632$\\$\pm 0.0007$} & \shortstack{$0.7286$\\$\pm 0.0003$} \\
\addlinespace[3pt]
GuidanceNFT & \shortstack{$\mathbf{0.7505}$\\$\pm 0.0007$} & \shortstack{$\mathbf{0.7667}$\\$\pm 0.0006$} & \shortstack{$\mathbf{0.7842}$\\$\pm 0.0008$} & \shortstack{$\mathbf{0.7686}$\\$\pm 0.0002$} & \shortstack{$\mathbf{0.7822}$\\$\pm 0.0006$} & \shortstack{$\mathbf{0.7958}$\\$\pm 0.0004$} \\
\bottomrule
\end{tabular}
\end{table}

\subsection{\texorpdfstring{\rev{Training Cost}}{Training Cost}}
\label{app:training-cost}

Table~\ref{tab:problem2-full-iteration-cost} measures intervals between
consecutive completed policy updates, including rollout and reward
scoring. Median GenEval iteration time is 387.9 seconds for NFT and
672.4 seconds for GuidanceNFT, a ratio of 1.73; the Learned3 ratio is 1.77.
Known evaluation, checkpoint, and restart boundaries are excluded;
unmarked pauses remain. Component medians need
not sum to the median total.

\input{tables/problem2_timing}

The center estimator
is discarded at deployment, adding no inference passes.

{\revon\paragraph{Where the extra time goes.}
On GenEval, GuidanceNFT adds 284.5 seconds per iteration relative to NFT.
Center fitting accounts for 150.6 seconds (53\%). Generating and scoring
CFG-4 rollouts, which require conditional and empty-condition passes,
accounts for 68.7 seconds (24\%), and the policy update, which also
evaluates the frozen center predictor, for 65.4 seconds (23\%). The latter
two costs follow from training on guided rollouts; removing center fitting
entirely would leave a ratio of about 1.35. We therefore test two ways to
reduce the fitting cost (Table~\ref{tab:center-cost-ablation}).

\paragraph{Less frequent center refresh.}
On SD3.5-M with PickScore, we vary how often the center predictor is
refit and for how many epochs. Refitting every ten updates with ten epochs
uses the same fitting computation as the default schedule and gives the
highest reward at update 50. Halving this computation or freezing the
center returns the reward to NFT's level, while
reducing the time ratio to 1.58 and 1.37.

\paragraph{Fewer fitting timesteps.}
On GenEval, fitting the center at five of the ten rollout timesteps,
rather than all ten, lowers the time ratio from 1.73 to 1.56. GenEval
reward with CFG 4 at update 100 is 0.913, compared with 0.905 for the
default and 0.682 for NFT. Fitting at a single timestep lowers the ratio
to 1.40 and reaches 0.940.
Within these tests, center-fitting computation can be reduced without
degrading reward when the center is refit on fewer timesteps or less often
with the same fitting computation; freezing or underfitting the center
removes the gain.\par}

{\revon\paragraph{Comparison at matched training time.}
Table~\ref{tab:matched-time} compares NFT at update 200 with GuidanceNFT
at matched training time. GuidanceNFT's training time includes rollout,
reward scoring, center fitting, and the policy update. On GenEval and OCR,
GuidanceNFT uses the single-timestep center fitting above
(Table~\ref{tab:center-cost-ablation}), giving update 143 on GenEval and
142 on OCR; on Learned3, it uses the default fitting at update 100.
Within the same training
time, GuidanceNFT with CFG 4 obtains higher GenEval and Learned3 rewards
than NFT at either evaluation scale and comparable OCR to NFT without
guidance, whereas NFT with CFG 4 degrades on all three objectives.\par}

\begin{table}[!ht]
\centering
\small
\setlength{\tabcolsep}{6pt}
\caption{{\revon\textbf{SD3.5-M at matched training time.} Train reward
with evaluation seed 42 and 40 sampling steps. NFT is evaluated at update
200; GuidanceNFT checkpoints (update in parentheses) match its training
time. GenEval and OCR use single-timestep center
fitting. Bold marks the best value per
objective.}}
\label{tab:matched-time}
\begin{tabular}{@{}lcccc@{}}
\toprule
Method & Eval CFG & GenEval & OCR & Learned3 \\
\midrule
NFT & 1 & 0.898 (200) & \textbf{0.972} (200) & 1.518 (200) \\
NFT & 4 & 0.667 (200) & 0.187 (200) & 1.495 (200) \\
GuidanceNFT & 4 & \textbf{0.950} (143) & 0.969 (142) & \textbf{1.523} (100) \\
\bottomrule
\end{tabular}
\end{table}

\begin{table}[!ht]
\centering
\small
\setlength{\tabcolsep}{5pt}
\caption{{\revon\textbf{Reducing center-fitting cost on SD3.5-M.}
All methods are evaluated with CFG 4; NFT is trained with CFG-1
rollouts. Seconds per update are medians, and time ratios divide them by
NFT on the same hardware. Top: PickScore/26 at update 50 on DrawBench
prompts, one H200; schedules are written as refit interval / center
optimizer steps. Bottom: GenEval reward at update 100, two H200 GPUs, with the
center fitted at ten, five, or one rollout timestep.}}
\label{tab:center-cost-ablation}
\begin{tabular}{@{}llccc@{}}
\toprule
Method & Center setting & Reward & Seconds / update & Time ratio \\
\midrule
\multicolumn{5}{@{}l}{\emph{PickScore/26, update 50 (refit interval / optimizer steps)}} \\
NFT & -- & 0.8580 & 182 & 1.00$\times$ \\
GuidanceNFT & 1 / 1 (default) & 0.8654 & 324 & 1.78$\times$ \\
GuidanceNFT & 10 / 10 & \textbf{0.8728} & 324 & 1.78$\times$ \\
GuidanceNFT & 10 / 5 & 0.8591 & 287 & 1.58$\times$ \\
GuidanceNFT & fixed & 0.8587 & 249 & 1.37$\times$ \\
\midrule
\multicolumn{5}{@{}l}{\emph{GenEval, update 100 (fitting timesteps)}} \\
NFT & -- & 0.6819 & 387.9 & 1.00$\times$ \\
GuidanceNFT & 10 (default) & 0.9054 & 672.4 & 1.73$\times$ \\
GuidanceNFT & 5 & 0.9132 & 605.8 & 1.56$\times$ \\
GuidanceNFT & 1 & \textbf{0.9396} & 541.2 & 1.40$\times$ \\
\bottomrule
\end{tabular}
\end{table}

\FloatBarrier
\section{SiT and Model Guidance Evaluations}
\label{app:sit-checkpoints}

Table~\ref{tab:intrinsic-guidance-results} reports fixed-checkpoint
PickScore for SiT and Model Guidance, complementing the training
trajectories in the main text.

\input{tables/model_guidance_results_main}

Each SiT checkpoint--CFG coordinate contains 6,000 images over three
evaluation seeds. Table~\ref{tab:sit-imagenet-checkpoints} reports raw
PickScore over the retained checkpoint sequence, complementing the
normalized trajectories in Figure~\ref{fig:explicit-guidance-diagnosis}(a).

\paragraph{Pretrained-empty SiT control.}
Replacing NFT's empty prediction with the pretrained one gives a modest
recovery but does not prevent its late decline. At update 400,
PickScore/26 is 0.7028 for NFT, 0.7101 with substitution, and 0.8299
for GuidanceNFT. NFT peaks at 0.7875 at update 200; GuidanceNFT improves through
update 400. This inference-only control uses CFG 1.5, the historical
three-channel SiT guidance, ten SDE steps, batch size eight, and the
same three evaluation seeds at updates 100--400 every 50 updates.
Source population SD is multiplied by $\sqrt{3/2}$ for sample SD.

\input{tables/sit_imagenet_checkpoint_results}

\paragraph{Selected PickScore examples.}
Figure~\ref{fig:mg-qualitative} compares NFT at update 200 with GuidanceNFT at
update 400. The left panel uses SiT with 250 sampling steps; the right uses
Model Guidance with 1,000 steps. SiT uses CFG 1 for NFT and CFG 1.5 for
GuidanceNFT; Model Guidance uses external CFG 1. Noise is paired
across methods within each column. \rev{PickScore for these checkpoints is
reported in Table~\ref{tab:intrinsic-guidance-results}.}

\subsection{\texorpdfstring{\rev{Reward-Tilted FID}}{Reward-Tilted FID}}
\label{app:tilted-fid}

\begin{figure}[!ht]
\centering
\includegraphics[width=\linewidth]{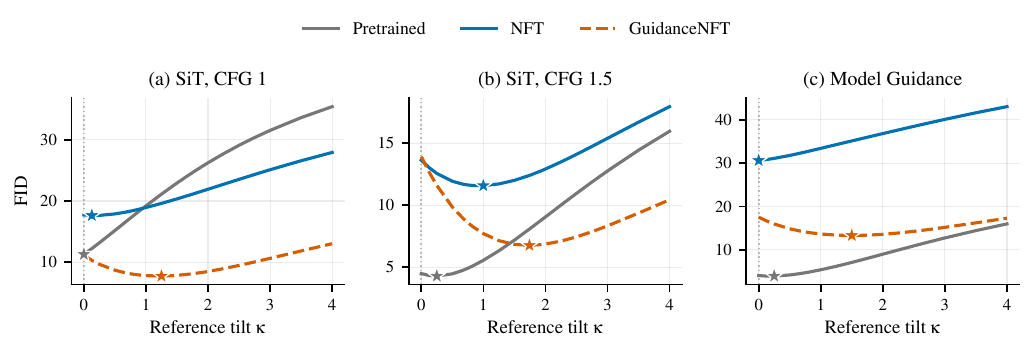}
\caption{{\revon\textbf{FID against reward-tilted ImageNet references at
update 200.} The reference weights each class's images by
$\exp(\kappa z)$, where $z$ is the within-class PickScore $z$-score;
$\kappa=0$ (dotted) gives standard FID. Stars mark each model's lowest FID,
reported as tFID in Tables~\ref{tab:imagenet-tilted-fid}
and~\ref{tab:imagenet-fid-full}.}}
\label{fig:imagenet-tilted-fid-curves}
\end{figure}

{\revon Standard FID measures distance to the ImageNet data distribution,
whereas reward fine-tuning is designed to move away from it toward higher
reward. We therefore also compare samples with a reward-tilted reference.
We score all 1,281,167 ImageNet training images with the same PickScore
model used for training and tilt the reference toward high reward:
\begin{equation}
p_\kappa(x\mid y)\propto p_{\rm data}(x\mid y)\,e^{\kappa z(x)},
\qquad
\mathrm{tFID}=\min_{\kappa\in\mathcal B}\mathrm{FID}\big(p_\theta,\,p_\kappa\big),
\label{eq:tilted-fid}
\end{equation}
where $z(x)$ is the PickScore $z$-score of $x$ within its class $y$ and each
class keeps equal weight. Setting $\kappa=0$ recovers standard FID.
Generated samples are 10,000 images, ten per class, from a 250-step
Euler--Maruyama sampler with shared noise across methods.
NFT and GuidanceNFT are compared at update 200, where NFT's reward
peaks (Table~\ref{tab:checkpoint-selection}), so that NFT is evaluated
before its degradation.
The set $\mathcal B$ is shared by all models and contains only non-negative
tilts: a grid from 0 to 4 together with, for each model, the $\kappa\ge 0$
whose tilted reference matches that model's mean PickScore, when it exists.

Figure~\ref{fig:imagenet-tilted-fid-curves} shows FID across tilt
strengths. The pretrained models are closest to the untilted or mildly
tilted reference, whereas GuidanceNFT reaches its lowest FID at larger
$\kappa$, consistent with samples that resemble higher-reward real images.
NFT's curves remain above GuidanceNFT's for all $\kappa>0$ in every
setting.

Table~\ref{tab:imagenet-fid-full} also reports standard FID ($\kappa=0$).
At update 200, GuidanceNFT
has lower standard FID than NFT on
SiT with CFG 1 (11.11 versus 17.72) and on Model Guidance (17.52 versus
30.61), and comparable standard FID on SiT with CFG 1.5 (13.93 versus
13.64).\par}

\input{tables/imagenet_fid_full}

\FloatBarrier
\section{ELF-B Guidance Sweep Measurements}
\label{app:elf-sweep-measurements}

Table~\ref{tab:elf-main-sweep-coordinates} reports every measured point
in Figure~\ref{fig:sccfg-main-sweep}, including the extended guidance
weights. The evaluation protocol is given in
Appendix~\ref{app:experimental-protocol}. These sweep measurements are
separate from the 1,000-sample repetition diagnostic shown in the
figure's text examples.

\input{tables/elf_main_sweep_coordinates}
\FloatBarrier

\section{GSM8K Reasoning Protocols}
\label{app:qwen-sccfg-online}

The reference checkpoints are from the ELF-B reasoning model used in
CEDR~\citep{anonymous2026lfrm}, built on ELF~\citep{hu2026elf}. It uses
Qwen-derived feature groups and a learned prompt encoder; inference
does not execute the Qwen teacher Transformer.

\paragraph{CEDR's NFT implementation.}
\label{app:lfrm-nft-implementation}
CEDR's adaptation uses the following detached correction for
current, old, and fixed-reference roles $p$:
\begin{equation}
\widetilde v_p=v_p^{\rm main}
-\operatorname{sg}\!\left[b_{\rm sc}(1-g^{-1})
 (v_p^{\rm main}-v_p^0)\right],\qquad b_{\rm sc}\in\{0,1\}.
\label{eq:qwen-reconstructed-center}
\end{equation}
Here $v_p^0$ uses zero answer self-conditioning, while the main pass uses
the detached bootstrap prediction when $b_{\rm sc}=1$; both retain the
question prompt. The loss samples $b_{\rm sc}\sim\mathrm{Bernoulli}(0.5)$
and $\log(1+g)\sim\mathrm{Uniform}(\log 1.5,\log 6)$, independently of
the SCCFG-2 rollout scale. The correction retains the current main-pass
Jacobian. The corrected fields enter NFT with $\beta=1$, NFT coefficient
5, and a squared-velocity reference penalty with coefficient 0.1,
averaged over latent coordinates and valid response tokens. Section 5.1
and Appendix E.2 of \citet{anonymous2026lfrm} give the full objective and
gradient routing. Our local native-field NFT and GuidanceNFT runs disable
this reconstruction.

\paragraph{Evaluation and training protocols.}
The main comparison evaluates all 1,319 GSM8K questions with 64 steps,
SCCFG 2, feature-group clocks $(2.5,2,1.5)$, and eight evaluation seeds.
Pretrained uses ELF/prompt EMA 0.9999/0.9999; all trained rows use
0.99/0.99. Table~\ref{tab:qwen-full1319-nft-comparison} reports
Pass@1/2/4/8 with problem-bootstrap std: 4,000 resamples of the 1,319
questions with replacement, preserving all
eight answers within each sampled question. Each replicate averages
unbiased question-level Pass@$k$ estimates; std uses the sample denominator.
All rows are recomputed from the saved answer cohorts.

Our NFT and GuidanceNFT runs use 32-step SCCFG-2 training rollouts and native
main-pass velocities, with
\texttt{training\_velocity\_reconstruction=false}.
The CEDR reference retains the question prompt when zeroing answer
self-conditioning. Its training also includes one reference solution per
15 generated answers, removes all-correct groups, and draws eight
logit-normal times per endpoint. Our runs use generated endpoints only,
retain groups containing both correct and incorrect answers, and use
31 draws from rollout time steps.
Pretrained is the released pre-NFT CEDR checkpoint.

Under the main protocol, NFT's Pass@1 is 40.03, 40.59, 41.21, 41.68,
39.63, 27.42, 17.48, and 12.23\% at updates 50--400, respectively.
The main table reports its best checkpoint (update 200), so GuidanceNFT's
gains are measured against NFT before its late degradation.

\paragraph{Ordinary SCCFG-2 checkpoint sweep.}
The sweep in Figure~\ref{fig:intrinsic-guidance-dynamics}(c) and
Table~\ref{tab:qwen-regular-sccfg2-sweep} evaluates the same runs with
32 steps, synchronous clocks $(1,1,1)$, and three seeds.
NFT reaches 36.57\% at update 200 before falling to 13.65\% at update
400. GuidanceNFT reaches 35.38\% at update 450 and 35.28\% at update 500.
GuidanceNFT achieves higher accuracy under the main asynchronous evaluation
and sustains performance through later training. NFT stops after its
training rollouts produce only incorrect answers; missing updates
450/500 denote a stopped run. Checkpoint selection follows the
conventions in Appendix~\ref{app:experimental-protocol}.

\input{tables/problem2_qwen_ordinary_sweep}

\FloatBarrier
\section{\texorpdfstring{\rev{Unguided Rollouts with a Trained Empty Branch}}{Unguided Rollouts with a Trained Empty Branch}}
\label{app:unguided-alternatives}

{\revon A simpler route to a reward-tuned guided model keeps unguided
rollouts, for which NFT's standard center is matched, and additionally
trains the empty-condition branch so that guidance can be applied at
inference. We test this approach with \emph{Dropout NFT}, which
replaces the condition with the empty condition in the NFT objective with
probability $p$. It uses unguided rollouts during training and guidance at
evaluation.

\paragraph{SiT-XL/2.}
On ImageNet with PickScore, we train Dropout NFT ($p=0.1$) for 200 updates with CFG-1 rollouts and
evaluate it with the protocol of Table~\ref{tab:imagenet-fid-full}
(Table~\ref{tab:unguided-dropout-sit}). At update 200, where NFT's reward
peaks, Dropout NFT has slightly lower PickScore than NFT at both
evaluation scales, lower IS, and higher standard and tilted FID.
GuidanceNFT is best on every metric except standard FID with CFG 1.5,
where it is comparable to NFT.

\begin{table}[!ht]
\centering
\small
\setlength{\tabcolsep}{4pt}
\caption{{\revon\textbf{Dropout NFT on SiT-XL/2 at update 200.}
10,000 samples (ten per class, 250 SDE steps); FID is standard FID
($\kappa=0$) and tFID the reward-tilted FID of
Appendix~\ref{app:tilted-fid}. NFT and GuidanceNFT rows repeat
Table~\ref{tab:imagenet-fid-full}. Dropout NFT uses $p=0.1$ and CFG-1
rollouts. Bold marks the best fine-tuned value.}}
\label{tab:unguided-dropout-sit}
\begin{tabular}{@{}lcccccccc@{}}
\toprule
& \multicolumn{4}{c}{Eval CFG 1} & \multicolumn{4}{c}{Eval CFG 1.5} \\
\cmidrule(lr){2-5}\cmidrule(l){6-9}
Method & PickScore$\uparrow$ & IS$\uparrow$ & FID$\downarrow$ & tFID$\downarrow$
 & PickScore$\uparrow$ & IS$\uparrow$ & FID$\downarrow$ & tFID$\downarrow$ \\
\midrule
NFT & 20.398 & 111.0 & 17.72 & 17.64 & 20.645 & 181.9 & \textbf{13.64} & 11.58 \\
Dropout NFT & 20.372 & 105.1 & 18.99 & 18.95 & 20.630 & 172.7 & 14.15 & 12.28 \\
GuidanceNFT & \textbf{20.907} & \textbf{200.4} & \textbf{11.11} & \textbf{7.73} & \textbf{21.113} & \textbf{270.1} & 13.93 & \textbf{6.76} \\
\bottomrule
\end{tabular}
\end{table}

\paragraph{ELF-B.}
We further test this variant on ELF-B with unguided rollouts. With a
dropout rate of 0.5, training collapses into repetitive text, with
12\%--16\% of the word 4-grams in each sample being repeats; with a rate of
0.1, it trails GuidanceNFT on both matched-entropy PPL and matched-PPL
entropy (Table~\ref{tab:unguided-dropout-elf} and
Figure~\ref{fig:unguided-alternatives-elf}).

\begin{table}[!ht]
\centering
\small
\setlength{\tabcolsep}{6pt}
\caption{{\revon\textbf{Dropout NFT on ELF-B (OpenWebText).}
Perplexity at unigram entropy 5.15 and entropy at perplexity 17,
interpolated along each perplexity--entropy curve after 100 updates.
Dropout rates are given in parentheses; ``not reached'' means the curve
does not attain the target value.}}
\label{tab:unguided-dropout-elf}
\begin{tabular}{@{}llcc@{}}
\toprule
Method & Training rollout & PPL at entropy 5.15 $\downarrow$ & Entropy at PPL 17 $\uparrow$ \\
\midrule
Pretrained & -- & 24.0 & not reached \\
NFT & SCCFG 1 & 18.2 & 5.134 \\
Dropout NFT (0.1) & SCCFG 1 & 20.5 & 5.115 \\
Dropout NFT (0.5) & SCCFG 1 & not reached & not reached \\
GuidanceNFT & SCCFG 3 & \textbf{16.1} & \textbf{5.176} \\
\bottomrule
\end{tabular}
\end{table}

\begin{figure}[!ht]
\centering
\includegraphics[width=0.6\linewidth]{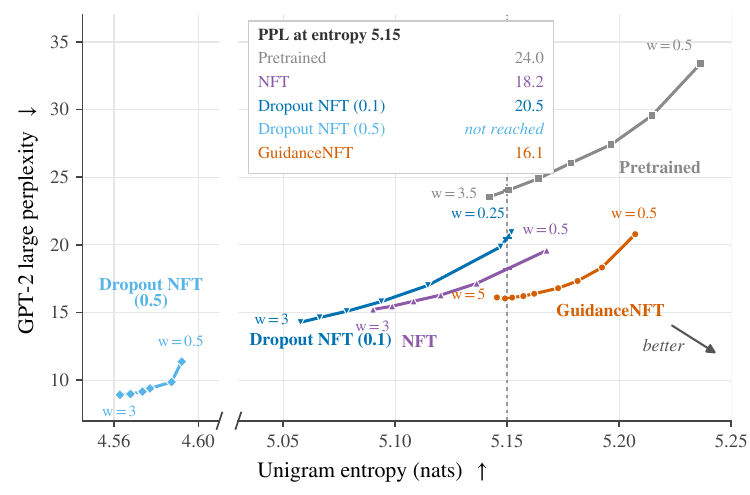}
\caption{{\revon\textbf{Unguided-rollout alternatives on ELF-B.}
Perplexity--entropy curves over the evaluation guidance weight $w$ after
100 updates; lower right is better. The inset reports perplexity at
entropy 5.15. The Dropout NFT (0.5) curve lies below entropy 4.6 and is
shown on a broken axis.}}
\label{fig:unguided-alternatives-elf}
\end{figure}

\paragraph{SD3.5-M.}
With the OCR
objective at update 200 (Table~\ref{tab:unguided-dropout-ocr}), Dropout
NFT partially mitigates the collapse of NFT with CFG 4 (OCR 0.724 versus
0.230), but its OCR reward is lower than that of NFT with CFG 1 and much
lower than that of GuidanceNFT. Its held-out metrics are mixed relative to
NFT with CFG 4, whereas GuidanceNFT is best on every metric.\par}

\begin{table}[!ht]
\centering
\small
\setlength{\tabcolsep}{4pt}
\caption{{\revon\textbf{Dropout NFT on SD3.5-M, OCR objective, update 200.}
All methods are trained on OCR and evaluated with seed 42 and 40 sampling
steps on the held-out prompts of Table~\ref{tab:sd35-full-cfg-update200}.
Dropout NFT uses $p=0.1$. Bold marks the best value per column; underline
marks the second best.}}
\label{tab:unguided-dropout-ocr}
\begin{tabular}{@{}lccccccccc@{}}
\toprule
Method & Eval CFG & OCR & GenEval & PickScore & CLIPScore & HPSv2.1 & Aesthetic & ImgRwd \\
\midrule
NFT & 1 & \underline{0.971} & 0.032 & 19.60 & 0.207 & 0.137 & 4.21 & $-1.69$ \\
NFT & 4 & 0.230 & 0.261 & 20.23 & 0.242 & \underline{0.188} & \underline{4.60} & \underline{$-0.87$} \\
Dropout NFT & 4 & 0.724 & \underline{0.319} & \underline{20.36} & \underline{0.249} & 0.183 & 4.46 & $-0.91$ \\
GuidanceNFT & 4 & \textbf{0.975} & \textbf{0.496} & \textbf{21.81} & \textbf{0.274} & \textbf{0.254} & \textbf{5.11} & $\mathbf{+0.40}$ \\
\bottomrule
\end{tabular}
\end{table}

%% file: tables/problem2_checkpoint_selection.tex
\begin{table}[!ht]
\centering\small
\setlength{\tabcolsep}{4pt}
\caption{\textbf{Checkpoints used for the main comparisons.}
NFT is selected by PickScore for SiT/MG and by Pass@1 under the
64-step asynchronous protocol for GSM8K. GuidanceNFT uses the reported
fixed checkpoints.}
\label{tab:checkpoint-selection}
\begin{tabular}{@{}lp{0.33\linewidth}cc@{}}
\toprule
Setting & Retained NFT candidates & NFT selected & GuidanceNFT reported \\
\midrule
SiT / ImageNet & 100--400, every 50 updates & 200 & 400 \\
Model Guidance / ImageNet & 200, 400 & 200 & 400 \\
Qwen / GSM8K & 50--400, every 50 updates & 200 & 500 \\
\bottomrule
\end{tabular}
\end{table}

%% file: tables/revision_fixed_checkpoint_ci.tex
\begin{table}[!ht]
\centering\small
\setlength{\tabcolsep}{4pt}
\caption{\textbf{SD3.5 at update 200.} Stds describe paired differences across 553 prompt means for GenEval dense reward, six task means for GenEval task-macro success, and 200 prompt means for Learned3. Each GenEval prompt retains four images; each Learned3 prompt retains ten images across two evaluation seeds. Learned3 is PickScore/26 + CLIPScore + HPSv2.1.}
\label{tab:revision-fixed-ci}
\begin{tabular}{@{}llr@{}}
\toprule
Metric & Comparator & GuidanceNFT $-$ comparator (mean $\pm$ std) \\
\midrule
GenEval dense & NFT, CFG 4 & $+0.2462 \pm 0.3182$ \\
GenEval dense & NFT, CFG 1 & $+0.0143 \pm 0.1737$ \\
GenEval dense & NFT + pretrained empty branch, CFG 4 & $+0.1043 \pm 0.2719$ \\
GenEval task macro & NFT, CFG 4 & $+0.3053 \pm 0.1353$ \\
GenEval task macro & NFT, CFG 1 & $+0.0269 \pm 0.0256$ \\
GenEval task macro & NFT + pretrained empty branch, CFG 4 & $+0.1337 \pm 0.0406$ \\
Learned3 sum & NFT, CFG 4 & $+0.0176 \pm 0.0357$ \\
Learned3 sum & NFT, CFG 1 & $-0.0019 \pm 0.0398$ \\
Learned3 sum & NFT + pretrained empty branch, CFG 4 & $-0.0049 \pm 0.0408$ \\
\bottomrule
\end{tabular}
\end{table}

%% file: tables/sd35_table1_seed_sd.tex
\begin{table}[!ht]
\centering\footnotesize
\setlength{\tabcolsep}{1.5pt}
\caption{\textbf{SD3.5 rewards and evaluation-seed variation.} Each cell shows the mean on the first line and $\pm$ sample std on the second, using two evaluation seeds per fixed update-200 checkpoint. Train reward uses on-domain test evaluations; the remaining metrics follow Table~\ref{tab:sd35-cfg4-update200}. All evaluations use 40 sampling steps at $512\times512$; PickScore is unscaled. Bold marks the best mean within each objective.}
\label{tab:sd35-table1-seed-sd}
\label{tab:sd35-drawbench-seed-sd}
\begin{tabular*}{\linewidth}{@{\extracolsep{\fill}}ll*{8}{c}@{}}
\toprule
 & & Train & & & & & & & \\
Objective & Method & reward & GenEval & OCR & PickScore & CLIPScore & HPSv2.1 & Aesthetic & ImgRwd \\
\midrule
GenEval & NFT {\scriptsize w/o CFG} & 0.900 & 0.855 & 0.434 & 21.70 & 0.265 & 0.234 & 4.886 & 0.271 \\
 & & $\pm 0.0024$ & $\pm 0.0077$ & $\pm 0.0059$ & $\pm 0.0021$ & $\pm 0.0002$ & $\pm 0.0001$ & $\pm 0.0132$ & $\pm 0.0178$ \\[2pt]
 & NFT {\scriptsize w/ CFG} & 0.657 & 0.565 & 0.624 & 21.19 & 0.271 & 0.216 & 4.773 & 0.148 \\
 & & $\pm 0.0139$ & $\pm 0.0003$ & $\pm 0.0144$ & $\pm 0.0061$ & $\pm 0.0014$ & $\pm 0.0006$ & $\pm 0.0116$ & $\pm 0.0224$ \\[2pt]
 & GuidanceNFT & \textbf{0.914} & \textbf{0.877} & \textbf{0.674} & \textbf{21.83} & \textbf{0.277} & \textbf{0.250} & \textbf{4.947} & \textbf{0.682} \\
 & & $\pm 0.0015$ & $\pm 0.0013$ & $\pm 0.0016$ & $\pm 0.0099$ & $\pm 0.0004$ & $\pm\,{<}0.0001$ & $\pm 0.0128$ & $\pm 0.0153$ \\
\midrule
OCR & NFT {\scriptsize w/o CFG} & 0.971 & 0.035 & 0.971 & 19.61 & 0.208 & 0.137 & 4.208 & $-$1.679 \\
 & & $\pm 0.0014$ & $\pm 0.0042$ & $\pm 0.0008$ & $\pm 0.0118$ & $\pm 0.0009$ & $\pm 0.0006$ & $\pm 0.0036$ & $\pm 0.0183$ \\[2pt]
 & NFT {\scriptsize w/ CFG} & 0.192 & 0.265 & 0.235 & 20.21 & 0.242 & 0.188 & 4.596 & $-$0.919 \\
 & & $\pm 0.0067$ & $\pm 0.0064$ & $\pm 0.0070$ & $\pm 0.0188$ & $\pm 0.0001$ & $\pm 0.0008$ & $\pm 0.0051$ & $\pm 0.0642$ \\[2pt]
 & GuidanceNFT & \textbf{0.975} & \textbf{0.488} & \textbf{0.974} & \textbf{21.83} & \textbf{0.274} & \textbf{0.256} & \textbf{5.120} & \textbf{0.425} \\
 & & $\pm 0.0010$ & $\pm 0.0099$ & $\pm 0.0009$ & $\pm 0.0293$ & $\pm 0.0003$ & $\pm 0.0022$ & $\pm 0.0072$ & $\pm 0.0426$ \\
\midrule
Learned3 & NFT {\scriptsize w/o CFG} & 1.519 & 0.671 & 0.497 & \textbf{23.51} & 0.284 & \textbf{0.332} & 5.606 & 1.397 \\
 & & $\pm 0.0003$ & $\pm 0.0061$ & $\pm 0.0075$ & $\pm 0.0150$ & $\pm 0.0001$ & $\pm 0.0008$ & $\pm 0.0135$ & $\pm 0.0092$ \\[2pt]
 & NFT {\scriptsize w/ CFG} & 1.495 & \textbf{0.701} & 0.579 & 22.99 & 0.290 & 0.326 & 5.397 & 1.414 \\
 & & $\pm 0.0002$ & $\pm 0.0026$ & $\pm 0.0001$ & $\pm 0.0109$ & $\pm 0.0001$ & $\pm 0.0002$ & $\pm 0.0034$ & $\pm 0.0005$ \\[2pt]
 & GuidanceNFT & \textbf{1.528} & 0.630 & \textbf{0.692} & 23.24 & \textbf{0.292} & \textbf{0.332} & \textbf{5.653} & \textbf{1.448} \\
 & & $\pm 0.0003$ & $\pm 0.0003$ & $\pm 0.0011$ & $\pm 0.0079$ & $\pm 0.0002$ & $\pm 0.0001$ & $\pm 0.0003$ & $\pm 0.0020$ \\
\bottomrule
\end{tabular*}
\end{table}

%% file: tables/problem2_geneval_tasks.tex
\begin{table}[!ht]
\centering\small
\setlength{\tabcolsep}{4pt}
\caption{\textbf{GenEval success by task.} Results use update 200, evaluation CFG 4, one evaluation seed, and four images per prompt. Differences and sample stds are in percentage points across paired prompt means within each task.}
\label{tab:problem2-geneval-tasks}
\begin{tabular}{@{}lrrrr@{}}
\toprule
Task & Prompts & NFT (\%) & GuidanceNFT (\%) & Difference $\pm$ std \\
\midrule
color\_attr & 100 & 22.25 & 74.25 & $+52.00 \pm 39.99$ \\
colors & 94 & 61.70 & 82.98 & $+21.28 \pm 39.99$ \\
counting & 80 & 58.75 & 92.19 & $+33.44 \pm 39.17$ \\
position & 100 & 56.50 & 87.00 & $+30.50 \pm 46.00$ \\
single\_object & 80 & 87.19 & 99.06 & $+11.88 \pm 26.38$ \\
two\_object & 99 & 59.34 & 93.43 & $+34.09 \pm 40.00$ \\
\bottomrule
\end{tabular}
\end{table}

%% file: tables/problem2_timing.tex
\begin{table}[!ht]
\centering\small
\setlength{\tabcolsep}{3pt}
\caption{\textbf{Recorded SD3.5 training costs.} Host seconds on two H200 GPUs, FP16, microbatch 24 per GPU, and 1,152 rollout endpoints per update. $n$ counts eligible complete iterations; IQR is the 25th--75th percentile. Ratios divide iteration medians by NFT within each task. The total includes rollout, reward scoring, center fitting, and policy updates; component medians need not sum to the total median. Known evaluation/checkpoint and restart boundaries are excluded. Unmarked long pauses remain (2/3 GenEval and 1/0 Learned3 intervals exceed five times the respective median for NFT/GuidanceNFT). Deployment requires no auxiliary-center pass.}
\label{tab:problem2-full-iteration-cost}
\begin{tabular}{@{}llrrrrrr@{}}
\toprule
Task & Method & $n$ & \shortstack{Rollout/\\reward} & Center & Policy & \shortstack{Iteration\\median [IQR]} & Ratio \\
\midrule
GenEval & NFT & 193 & 113.6 & 0.0 & 274.3 & 387.9 [387.2, 388.7] & 1.00$\times$ \\
GenEval & GuidanceNFT & 184 & 182.2 & 150.6 & 339.7 & 672.4 [671.1, 674.3] & 1.73$\times$ \\
\midrule
Learned3 & NFT & 94 & 92.5 & 0.0 & 274.5 & 367.3 [366.7, 367.8] & 1.00$\times$ \\
Learned3 & GuidanceNFT & 96 & 160.5 & 150.5 & 339.2 & 650.2 [649.9, 650.6] & 1.77$\times$ \\
\bottomrule
\end{tabular}
\end{table}

%% file: tables/model_guidance_results_main.tex
\begin{table}[!ht]
\centering\small
\setlength{\tabcolsep}{5pt}
\caption{Raw PickScore ($\uparrow$), mean $\pm$ sample std. Each checkpoint uses three evaluation seeds and 6,000 images; standard deviations use the sample denominator $n-1$. NFT uses its best observed checkpoint (200); GuidanceNFT uses 400. MG has external CFG 1. Bold indicates a larger point estimate.}
\label{tab:intrinsic-guidance-results}
\begin{tabular}{@{}lccc@{}}
\toprule
Method & SiT, CFG 1 & SiT, CFG 1.5 & Model Guidance \\
\midrule
Pretrained & 18.849 $\pm$ 0.012 & 19.287 $\pm$ 0.015 & 19.239 $\pm$ 0.003 \\
NFT & 20.161 $\pm$ 0.019 & 20.476 $\pm$ 0.025 & 20.010 $\pm$ 0.023 \\
GuidanceNFT & \textbf{21.465 $\pm$ 0.006} & \textbf{21.579 $\pm$ 0.013} & \textbf{21.580 $\pm$ 0.018} \\
\bottomrule
\end{tabular}
\end{table}

%% file: tables/sit_imagenet_checkpoint_results.tex
\begin{table}[!ht]
\centering\small
\setlength{\tabcolsep}{2pt}
\renewcommand{\arraystretch}{1.12}
\caption{\textbf{SiT-XL/2 unscaled PickScore ($\uparrow$) at every retained checkpoint} (columns 100--400). Each cell shows the mean with $\pm$ sample std beneath it across three evaluation-seed means, conditional on a fixed checkpoint. Each seed covers all 1,000 ImageNet classes with two images per class (2,000 images per seed; 6,000 images per entry), using ten-step SDE sampling. NFT trains with CFG-1 rollouts. GuidanceNFT trains with CFG-1.5 rollouts. Bold marks the best checkpoint within each method and evaluation-CFG row.}
\label{tab:sit-imagenet-checkpoints}
\begin{tabular*}{\linewidth}{@{\extracolsep{\fill}}lc*{7}{c}@{}}
\toprule
Method & CFG & 100 & 150 & 200 & 250 & 300 & 350 & 400 \\
\midrule
NFT & 1 & \shortstack{$19.392$\\$\pm 0.025$} & \shortstack{$19.791$\\$\pm 0.007$} & \shortstack{$\mathbf{20.161}$\\$\pm 0.019$} & \shortstack{$19.419$\\$\pm 0.023$} & \shortstack{$19.013$\\$\pm 0.009$} & \shortstack{$18.477$\\$\pm 0.003$} & \shortstack{$18.317$\\$\pm 0.003$} \\
\addlinespace[3pt]
 & 1.5 & \shortstack{$19.814$\\$\pm 0.006$} & \shortstack{$20.179$\\$\pm 0.010$} & \shortstack{$\mathbf{20.476}$\\$\pm 0.025$} & \shortstack{$19.747$\\$\pm 0.026$} & \shortstack{$19.094$\\$\pm 0.006$} & \shortstack{$18.390$\\$\pm 0.003$} & \shortstack{$18.273$\\$\pm 0.004$} \\
\midrule
GuidanceNFT & 1 & \shortstack{$19.442$\\$\pm 0.016$} & \shortstack{$19.880$\\$\pm 0.015$} & \shortstack{$20.375$\\$\pm 0.016$} & \shortstack{$20.853$\\$\pm 0.003$} & \shortstack{$21.157$\\$\pm 0.009$} & \shortstack{$21.345$\\$\pm 0.008$} & \shortstack{$\mathbf{21.465}$\\$\pm 0.006$} \\
\addlinespace[3pt]
 & 1.5 & \shortstack{$19.924$\\$\pm 0.005$} & \shortstack{$20.319$\\$\pm 0.003$} & \shortstack{$20.728$\\$\pm 0.008$} & \shortstack{$21.097$\\$\pm 0.003$} & \shortstack{$21.325$\\$\pm 0.007$} & \shortstack{$21.474$\\$\pm 0.016$} & \shortstack{$\mathbf{21.579}$\\$\pm 0.013$} \\
\bottomrule
\end{tabular*}
\end{table}

%% file: tables/imagenet_fid_full.tex
\begin{table}[!ht]
\centering
\small
\setlength{\tabcolsep}{5pt}
\caption{{\revon\textbf{ImageNet FID and Inception Score at update 200.} Each row uses
10,000 samples (ten per class, 250 SDE steps). FID uses the class-balanced
ImageNet training set ($\kappa=0$); tFID is the lowest FID over the shared
set of non-negative tilt strengths, attained at $\hat\kappa$. IS is mean $\pm$ std over
ten splits.}}
\label{tab:imagenet-fid-full}
\begin{tabular}{@{}llccccc@{}}
\toprule
Setting & Method & PickScore$\uparrow$ & IS$\uparrow$ & FID$\downarrow$ & tFID$\downarrow$ & $\hat\kappa$ \\
\midrule
\multirow{3}{*}{SiT, CFG 1}
 & Pretrained & 19.885 & 115.3 $\pm$ 4.1 & 11.28 & 11.28 & 0.00 \\
 & NFT & 20.398 & 111.0 $\pm$ 5.6 & 17.72 & 17.64 & 0.13 \\
 & GuidanceNFT & 20.907 & 200.4 $\pm$ 8.8 & 11.11 & 7.73 & 1.25 \\
\midrule
\multirow{3}{*}{SiT, CFG 1.5}
 & Pretrained & 20.321 & 218.7 $\pm$ 6.3 & 4.48 & 4.28 & 0.25 \\
 & NFT & 20.645 & 181.9 $\pm$ 5.4 & 13.64 & 11.58 & 1.00 \\
 & GuidanceNFT & 21.113 & 270.1 $\pm$ 5.2 & 13.93 & 6.76 & 1.74 \\
\midrule
\multirow{3}{*}{Model Guidance}
 & Pretrained & 20.338 & 240.6 $\pm$ 7.2 & 4.05 & 3.93 & 0.25 \\
 & NFT & 20.335 & 94.8 $\pm$ 3.8 & 30.61 & 30.61 & 0.00 \\
 & GuidanceNFT & 21.017 & 231.4 $\pm$ 5.3 & 17.52 & 13.30 & 1.50 \\
\bottomrule
\end{tabular}
\end{table}

%% file: tables/elf_main_sweep_coordinates.tex
\begin{table}[!htbp]
\centering\small
\setlength{\tabcolsep}{4pt}
\begin{tabular*}{\linewidth}{@{\extracolsep{\fill}}c*{6}{r}@{}}
\toprule
\multirow{2}{*}{SCCFG $w$} & \multicolumn{2}{c}{Pretrained} & \multicolumn{2}{c}{CEDR-style} & \multicolumn{2}{c}{GuidanceNFT} \\
\cmidrule(lr){2-3}\cmidrule(lr){4-5}\cmidrule(lr){6-7}
& Entropy & PPL & Entropy & PPL & Entropy & PPL \\
\midrule
0.5 & 5.2363 & 33.3990 & 5.1210 & 19.6921 & 5.2071 & 20.7896 \\
1 & 5.2147 & 29.5693 & 5.1034 & 17.5670 & 5.1923 & 18.3294 \\
1.5 & 5.1964 & 27.4135 & 5.0902 & 16.6942 & 5.1814 & 17.3293 \\
2 & 5.1784 & 26.0441 & 5.0793 & 16.0856 & 5.1727 & 16.7972 \\
2.5 & 5.1640 & 24.9028 & 5.0705 & 15.6371 & 5.1621 & 16.3889 \\
3 & 5.1506 & 24.0447 & 5.0637 & 15.3213 & 5.1572 & 16.2141 \\
3.5 & 5.1421 & 23.5557 & & & 5.1522 & 16.1056 \\
4 & & & & & 5.1490 & 16.0393 \\
5 & & & & & 5.1454 & 16.1141 \\
\bottomrule
\end{tabular*}
\caption{Measured coordinates of the three ELF-B curves in Figure~\ref{fig:sccfg-main-sweep}. Entropy is in nats; PPL is evaluated by GPT-2 large. Each point uses 1,024 samples, a shared evaluation seed, 32 SDE steps, $\gamma=1.5$, and external CFG 1. CEDR-style and GuidanceNFT use update-100 checkpoints trained with SCCFG 1 and 3, respectively. Blank cells denote guidance weights not evaluated for that curve; all displayed numbers are measured values, without interpolation.}
\label{tab:elf-main-sweep-coordinates}
\end{table}

%% file: tables/problem2_qwen_ordinary_sweep.tex
\begin{table}[!ht]
\centering\small
\setlength{\tabcolsep}{2.5pt}
\caption{\textbf{Synchronous SCCFG-2 evaluation.} The sweep uses 1,319 GSM8K questions, 32 steps, synchronous clocks $(1,1,1)$, external CFG 1, ELF/prompt EMA $0.99/0.99$, and three evaluation seeds. Each cell shows mean accuracy (\%) with $\pm$ sample std beneath it, across evaluation seeds for one training run. Bold marks each method\textquotesingle s largest observed mean. NFT training stopped after its rollouts produced only incorrect answers, so update 400 is its last checkpoint (dashes). This protocol differs from the 64-step asynchronous main-table evaluation.}
\label{tab:qwen-regular-sccfg2-sweep}
\begin{tabular*}{\linewidth}{@{\extracolsep{\fill}}l*{10}{c}@{}}
\toprule
Update & 50 & 100 & 150 & 200 & 250 & 300 & 350 & 400 & 450 & 500 \\
\midrule
NFT & \shortstack{$32.12$\\$\pm 0.34$} & \shortstack{$34.22$\\$\pm 0.46$} & \shortstack{$34.77$\\$\pm 0.38$} & \shortstack{$\mathbf{36.57}$\\$\pm 0.31$} & \shortstack{$35.00$\\$\pm 0.88$} & \shortstack{$29.54$\\$\pm 1.18$} & \shortstack{$19.84$\\$\pm 1.35$} & \shortstack{$13.65$\\$\pm 0.35$} & --- & --- \\
\addlinespace[4pt]
GuidanceNFT & \shortstack{$31.39$\\$\pm 0.27$} & \shortstack{$31.99$\\$\pm 0.20$} & \shortstack{$33.21$\\$\pm 0.35$} & \shortstack{$34.37$\\$\pm 0.12$} & \shortstack{$34.14$\\$\pm 0.12$} & \shortstack{$35.08$\\$\pm 0.62$} & \shortstack{$34.60$\\$\pm 0.49$} & \shortstack{$35.18$\\$\pm 0.30$} & \shortstack{$\mathbf{35.38}$\\$\pm 0.24$} & \shortstack{$35.28$\\$\pm 0.29$} \\
\bottomrule
\end{tabular*}
\end{table}